\documentclass[10pt,twocolumn,letterpaper]{article}

\usepackage{cvpr}              

\usepackage{algorithm}
\usepackage{algpseudocode}
\usepackage{booktabs}
\usepackage{multirow}
\usepackage{graphicx}   
\usepackage{float}
\usepackage[accsupp]{axessibility}

\newcommand{\best}[1]{\textbf{#1}}
\newcommand{\second}[1]{\underline{#1}}
\usepackage{float}
\usepackage{dblfloatfix}

\definecolor{cvprblue}{rgb}{0.21,0.49,0.74}
\usepackage[pagebackref,breaklinks,colorlinks,allcolors=cvprblue]{hyperref}

\def\paperID{*****} 
\def\confName{CVPR}
\def\confYear{2026}

\title{Seeing Through the Tool: A Controlled Benchmark for Occlusion Robustness in Foundation Segmentation Models}

\author{
Nhan Ho$^{1}$\thanks{Equal contribution.} \quad
Luu Le$^{2}$\footnotemark[1] \quad
Thanh-Huy Nguyen$^{3}$ \quad
Thien Nguyen$^{3}$ \quad
Xiaofeng Liu$^{4}$ \quad
Ulas Bagci$^{5}$\thanks{Corresponding author.}\\
$^{1}$Stony Brook University, USA \quad
$^{2}$AIMA Research Lab \quad $^{3}$Carnegie Mellon University, USA \\
$^{4}$Yale University, USA \quad
$^{5}$Northwestern University, USA
}

\begin{document}
\maketitle
\begin{abstract}

Occlusion, where target structures are partially hidden by surgical instruments or overlapping tissues, remains a critical yet underexplored challenge for foundation segmentation models in clinical endoscopy. We introduce \textbf{OccSAM-Bench}, a benchmark designed to systematically evaluate SAM-family models under controlled, synthesized surgical occlusion. Our framework simulates two occlusion types (i.e., surgical tool overlay and cutout) across three calibrated severity levels on three public polyp datasets. We propose a novel three-region evaluation protocol that decomposes segmentation performance into full, visible-only, and invisible targets. This metric exposes behaviors that standard amodal evaluation obscures, revealing two distinct model archetypes: \textit{Occluder-Aware} models (SAM, SAM 2, SAM 3, MedSAM3), which prioritize visible tissue delineation and reject instruments, and \textit{Occluder-Agnostic} models (MedSAM, MedSAM2), which confidently predict into occluded regions. SAM-Med2D aligns with neither and underperforms across all conditions. Ultimately, our results demonstrate that occlusion robustness is not uniform across architectures, and model selection must be driven by specific clinical intent—whether prioritizing conservative visible-tissue segmentation or the amodal inference of hidden anatomy.
\end{abstract}    

\section{Introduction}
\label{sec:intro}

\begin{figure}[t]
\centering
\includegraphics[width=\columnwidth]{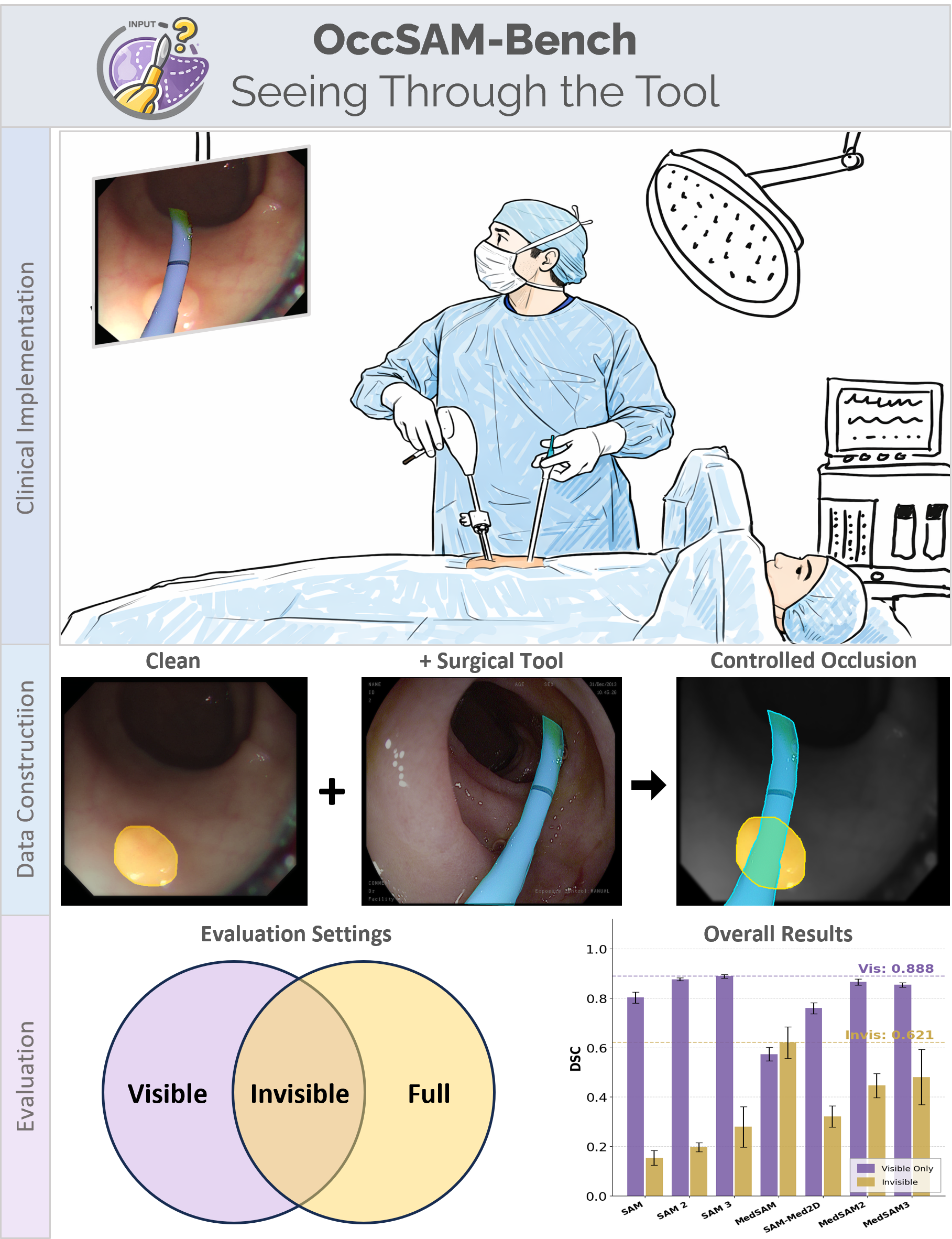}
\caption{\textbf{OccSAM-Bench: evaluating segmentation under surgical occlusion.}
Clinical endoscopy often involves partial occlusion of target anatomy by surgical instruments. OccSAM-Bench simulates such conditions by overlaying surgical tools onto polyp images to create controlled occlusions. Models are evaluated using a three-region protocol that separates \textbf{visible}, \textbf{invisible} (occluded), and \textbf{full} masks, revealing behaviors that standard full-mask evaluation alone cannot capture.}
\label{fig:teaser}
\end{figure}

Foundation models for image segmentation, led by the Segment Anything 
Model (SAM)~\cite{kirillov2023sam} and its successors SAM~2~\cite{ravi2024sam2} 
and SAM~3~\cite{carion2025sam3segmentconcepts}, have demonstrated remarkable 
zero-shot generalization capabilities. Recent medical adaptations, including 
MedSAM~\cite{ma2024medsam}, SAM-Med2D~\cite{cheng2023sammed2d}, 
MedSAM2~\cite{zhu2024medsam2}, and MedSAM3~\cite{liu2025medsam3delvingsegmentmedical} 
have further extended interactive segmentation paradigms into clinical settings. However,  
these models are almost exclusively benchmarked on \emph{clean}, highly curated medical 
images~\cite{huang2024samsurvey,mazurowski2023segment}. While recent studies in natural imaging have begun examining SAM's behavior under occlusion~\cite{sameo2025,stablesam2025}, the medical domain lacks corresponding scrutiny. This represents a significant vulnerability: in-real world clinical workflows, target structures are frequently obstructed by dynamic surgical instruments, posing a severe risk of downstream navigational or diagnostic errors if models fail to interpret these occlusions correctly. 

Although prior work such as SAMEO~\cite{sameo2025} explores the amodal segmentation of general objects , no existing benchmark systematically quantifies foundation model robustness against surgical-tool occlusion in endoscopy. Furthermore, standard evaluation paradigms rely heavily on full (amodal) mask overlap. This approach is fundamentally flawed in surgical contexts. A model that incorrectly hallucinates tissue over a surgical instrument may achieve a high full-mask score by coincidentally overlapping the hidden ground truth , while a model that correctly rejects the instrument is penalized. To address this, we introduce a benchmark that deliberately decomposes evaluation into visible, invisible, and full masks to align with distinct clinical interpretations.


\begin{figure*}[htbp]
\centering
\includegraphics[width=\textwidth]{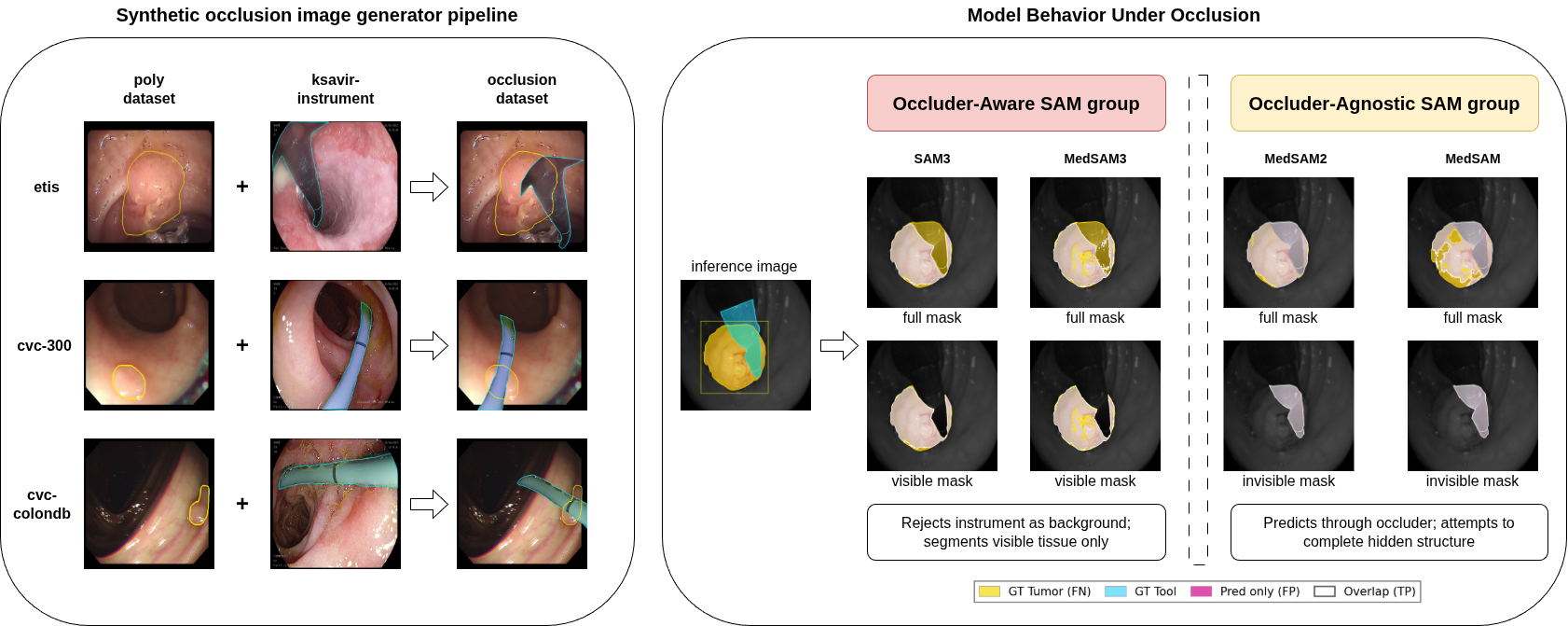}
\caption{\textbf{OccSAM-Bench overview.}
\textit{Left:} Surgical-tool occlusions are simulated by pasting instrument masks onto polyp images from three colonoscopy datasets.
\textit{Right:} SAM-family models exhibit two behaviors under occlusion:
\textbf{Occluder-Aware} models segment visible tissue while rejecting the instrument,
whereas \textbf{Occluder-Agnostic} models predict through the occluder.
Both can achieve similar full-mask scores, showing that standard evaluation obscures these differences.}
\label{fig:intro}
\end{figure*}

We introduce \textbf{OccSAM-Bench}, offering three primary contributions to the field of medical image segmentation:
\begin{itemize}
\item \textbf{Controlled Occlusion Framework:} We propose a principled, synthetic occlusion generation pipeline that injects surgical tool overlays and data cutouts at three calibrated severity levels, isolating the effects of visual confusion from missing data (Sec.~\ref{sec:occlusion_generation}).
\item \textbf{Three-Region Evaluation Protocol:} We develop a novel metric that decomposes performance into visible-only, invisible, and full mask targets (Sec.~\ref{sec:eval_protocol}). This directly addresses the shortcomings of standard amodal metrics by heavily penalizing clinically dangerous oversegmentation.
\item \textbf{Comprehensive Zero-Shot Benchmark:} We systematically evaluate seven state-of-the-art SAM-family models across three public polyp datasets (Sec.~\ref{sec:results}). Our analysis uncovers two distinct behavioral archetypes—Occluder-Aware and Occluder-Agnostic—proving that architectural priors and domain-specific fine-tuning dramatically alter how foundation models process partial visibility.
\end{itemize}


Our experiments reveal two consistent behavioral archetypes. 
\emph{Occluder-Aware} models (SAM, SAM~2, SAM~3, MedSAM3) reject the instrument as background and prioritize accurate delineation of visible tissue. \emph{Occluder-Agnostic} models (MedSAM, MedSAM2) predict into occluded regions, a tendency already visible at low and medium severity and suggestive of amodal completion behavior. Within this group, MedSAM2 stands 
out as the only model that maintains competitive visible DSC while achieving high invisible scores, likely due to its video-based fine-tuning strategy. SAM-Med2D aligns with neither pattern and underperforms across conditions. Together, these findings indicate that model selection under occlusion 
should be guided by clinical intent rather than clean-image performance alone (Sec.~\ref{sec:results}).
\section{Related Work}
\label{sec:related}
\textbf{Foundation Models in Medical Imaging.}
Limited annotation remains a major challenge in medical imaging, motivating the use of foundation models to better exploit unlabeled data through transferable priors and adaptable representations. Recent efforts have extended foundation-model-based learning beyond standard supervised settings, including unsupervised cryo-ET segmentation~\cite{uddin2025unsupervised}, SAM-based learning from unlabeled medical images~\cite{vu2026specialist}, robust adaptation in fetal ultrasound~\cite{le2026ultra}, and broader medical image understanding~\cite{xiao2025describe}. These advances are also related to semi-supervised and source-free segmentation under annotation scarcity and domain shift~\cite{nguyen2026adaptive,nguyen2026up2d}.

Segment Anything Model (SAM)~\cite{kirillov2023sam} introduced promptable segmentation trained on over one billion masks. SAM~2~\cite{ravi2024sam2} extended this to video with a dedicated occlusion head for handling object disappearance across frames, and SAM~3~\cite{carion2025sam3segmentconcepts} further improved spatial and concept-level understanding, trained on SA-Co, Meta's largest segmentation dataset to date. Medical adaptations include MedSAM~\cite{ma2024medsam} (full fine-tuning on 1.57M medical image-mask pairs across 11 modalities), SAM-Med2D~\cite{cheng2023sammed2d} (adapter modules with frozen SAM encoder), and MedSAM2~\cite{zhu2024medsam2} (SAM~2 fine-tuned treating medical images as video sequences). Prior evaluations~\cite{huang2024samsurvey,mazurowski2023segment,he2023sammedical} assess these on clean benchmarks, but no study examines occlusion robustness.

\textbf{Amodal Segmentation and Occlusion.}
Amodal segmentation, inference of complete object structures despite partial visibility, has been extensively explored in natural imaging paradigms~\cite{ke2021bcnet,qi2019amodal,follmann2019amodal,zhu2017cocoa}. 
Recent frameworks such as SAMEO~\cite{sameo2025} adapt SAM specifically for amodal mask decoding. However, directly transferring amodal evaluation metrics to surgical environments is problematic~\cite{ross2021smoke,rueckert2023smoke}. In endoscopy, instruments are explicitly non-target occluders; over-predicting target tissue into the spatial footprint of a surgical tool constitutes a hazardous false positive. Unlike SAMEO or natural scene benchmarks, our three-region protocol explicitly penalizes models that conflate foreign instruments with target anatomy, aligning the evaluation metric directly with surgical safety constraints. 

\section{Methodology}
\label{sec:method}

\subsection{Occlusion Generation}
\label{sec:occlusion_generation}

We implement two complementary occlusion types under a unified 
bin-controlled generation framework.

\noindent\textbf{Severity levels.}
We define three bins based on the fraction $r$ of target area occluded: 
\textbf{Low} (0--20\%), \textbf{Medium} (20--40\%), and \textbf{High} 
(40--60\%), where:
\begin{equation}
    r = \frac{|M_{\text{full}} \cap M_{\text{occluder}}|}{|M_{\text{full}}|}
\label{eq:occlusion_ratio}
\end{equation}
For each sample, a target bin is selected and a candidate occluder is 
generated. The overlap ratio is computed using Eq.~\ref{eq:occlusion_ratio}, 
and the sample is accepted only if $r$ falls within the specified bin. 
This rejection sampling loop runs for up to 50 attempts per image.

\noindent\textbf{Surgical tool paste.}
Binary masks of real surgical instruments are overlaid onto target images to partially obscure the anatomical structure. Tool instances are randomly sampled from the Kvasir-Instrument dataset library~\cite{jha2021kvasir_instrument}, scaled ($0.8\times$--$1.0\times$), and rotated $\pm 45^{\circ}$ to approximate dynamic surgical trajectories. While 2D overlay abstracts away the complex optical physics of in-vivo occlusions (e.g., localized tissue deformation and specular reflections), this synthetic injection is an indispensable experimental design choice. In genuine surgical recordings, the true extent of the occluded tissue is unknown, rendering exact amodal evaluation impossible. By synthesizing the occlusion over a known ground truth, our framework provides strict, mathematically verifiable bounds for the $M_{vis}$   and  $M_{inv}$ metrics, isolating the models' structural reasoning from pure guesswork.


\noindent\textbf{Cutout occlusion.}
Inspired by CutOut~\cite{devries2017cutout}, we remove image content within the target region. 
A rectangular mask with area proportional to the target size is placed near the target centroid with a bounded random offset. 
Its size is iteratively adjusted until the overlap ratio falls within the selected severity bin. 
Unlike tool paste, Cutout removes information without introducing foreign visual content, isolating the effect of missing data from visual confusion.

The dataset generator iterates over all source images to produce samples across severity bins, yielding roughly balanced distributions. 
The final dataset size depends on the source data and generation success rate.



\begin{algorithm}[htbp]
\caption{Controlled Surgical Tool Occlusion Generation}
\label{alg:generation}
\begin{algorithmic}[1]
\Require Image $I$, target mask $M$, bin $b$, tool library $\mathcal{T}$
\Ensure Occluded image $I'$, occlusion mask $M_o$, effective mask $M_e$
\State $[r_{\min}, r_{\max}] \leftarrow \text{BinBounds}(b)$
\For{attempt $= 1$ to $K=50$}
    \State $T \sim \text{Uniform}(\mathcal{T})$; \quad $s \sim \mathcal{U}(0.8, 1.0)$; \quad $\theta \sim \mathcal{U}(-45^{\circ}, 45^{\circ})$
    \State $T' \leftarrow \text{Rotate}(\text{Scale}(T, s), \theta)$
    \State $\mathbf{p} \leftarrow \text{Centroid}(M) + \boldsymbol{\delta}$
    \State $M_o \leftarrow \text{Paste}(T', \mathbf{p})$; \quad $r \leftarrow |M \cap M_o|/|M|$
    \If{$r_{\min} \leq r \leq r_{\max}$}
        \State $I' \leftarrow \text{Apply}(I, M_o)$; \quad $M_e \leftarrow M \setminus M_o$
        \Return $I', M_o, M_e$
    \EndIf
\EndFor
\end{algorithmic}
\end{algorithm}

\begin{algorithm}[htbp]
\caption{Controlled CutOut Generation}
\label{alg:cutout}
\begin{algorithmic}[1]
\Require Image $I$, tumor mask $M$, bin $b$, seed $s$
\Ensure Occluded image $I'$, occlusion mask $M_o$, effective mask $M_e$
\State $[r_{\min}, r_{\max}] \leftarrow \text{BinBounds}(b)$
\State $r^* \sim \mathcal{U}(r_{\min}, r_{\max})$
\State $[x_{\min}, x_{\max}, y_{\min}, y_{\max}] \leftarrow \text{BBox}(M)$
\State $A_M \leftarrow |M|$; \quad $A_t \leftarrow A_M \cdot r^*$; \quad $A_c \leftarrow A_t \cdot \mathcal{U}(1.2, 1.8)$
\State $\alpha \sim \mathcal{U}(0.5, 2.0)$; \quad $h \leftarrow \sqrt{A_c/\alpha}$; \quad $w \leftarrow A_c/h$
\State $\mathbf{p} \leftarrow \text{Centroid}(M) + \boldsymbol{\delta}$
\State $M_o \leftarrow \text{RectangleMask}(h, w, \mathbf{p})$
\State $r \leftarrow |M \cap M_o| / |M|$
\For{attempt $=1$ to $K=50$}
    \If{$r_{\min} < r \leq r_{\max}$}
        \State $I' \leftarrow \text{Apply}(I, M_o)$
        \State $M_e \leftarrow M \setminus M_o$
        \Return $I', M_o, M_e$
    \EndIf
    \If{$r \leq r_{\min}$}
        \State $h \leftarrow 1.3h$; \quad $w \leftarrow 1.3w$
    \Else
        \State $h \leftarrow 0.7h$; \quad $w \leftarrow 0.7w$
    \EndIf
    \State $M_o \leftarrow \text{RectangleMask}(h, w, \mathbf{p})$
    \State $r \leftarrow |M \cap M_o| / |M|$
\EndFor
\State $I' \leftarrow \text{Apply}(I, M_o)$
\State $M_e \leftarrow M \setminus M_o$
\Return $I', M_o, M_e$
\end{algorithmic}
\end{algorithm}

\subsection{Three-Region Evaluation Protocol}
\label{sec:eval_protocol}

Different evaluation targets reveal distinct model behaviors under occlusion. 
For each occluded sample, we define three complementary masks:
\begin{equation}
    M_{\text{vis}} = M_{\text{full}} \setminus M_{\text{occ}}, \quad
    M_{\text{inv}} = M_{\text{full}} \cap M_{\text{occ}} .
\end{equation}
Predictions are evaluated against all three:

\begin{enumerate}[leftmargin=*,itemsep=2pt]
\item \textbf{Visible-only mask} $M_{\text{vis}}$: The ground truth that is actually visible. 
It measures whether the model segments visible tissue while \emph{rejecting the occluder}. 
Predictions extending into the tool region are penalized as false positives.

\item \textbf{Invisible mask} $M_{\text{inv}}$: The hidden ground truth behind the occluder. 
High scores indicate predictions inside occluded regions. However, high invisible DSC does 
not necessarily imply true amodal reasoning, as it may arise from over-prediction or occluder confusion.

\item \textbf{Full mask} $M_{\text{full}}$: The complete target including occluded parts. 
While commonly used, this metric can be \emph{misleading}: predictions that incorrectly cover the 
instrument may still achieve high overlap.
\end{enumerate}

\noindent\textbf{Why full-mask evaluation is misleading.}
A prediction covering both the polyp and the overlaid tool yields high overlap with 
$M_{\text{full}}$ but low precision under $M_{\text{vis}}$. Thus, full-mask metrics may 
reward clinically incorrect predictions. We therefore recommend \textbf{visible-only DSC} 
as the primary robustness metric. The invisible score serves as a complementary probe of 
prediction beyond visible boundaries but should be interpreted cautiously.

\noindent\textbf{Prompt types.} We evaluate two prompt settings: 
(1)~\textbf{bounding box prompts} derived from the full ground-truth mask, and 
(2)~\textbf{single interior point prompts} sampled from pixels whose distance to the boundary 
exceeds the median distance. Other configurations (e.g., multiple clicks or positive/negative pairs) 
are not explored.

\noindent\textbf{Metrics.}
We report Dice Similarity Coefficient (DSC) and the 95th-percentile Hausdorff Distance (HD95).

\subsection{Prompt Generation Strategy}
\label{sec:prompting}

To ensure fair comparison across models, prompts are generated automatically from ground-truth annotations.

\noindent\textbf{Point prompts.}
Following~\cite{mazurowski2023segment}, we compute the Euclidean distance from each foreground pixel 
to the nearest boundary and sample a pixel with distance above the median, placing the point near the 
object center.

\noindent\textbf{Box prompts.}
We compute a tight bounding box $(x_{\min}, y_{\min}, x_{\max}, y_{\max})$ from the mask and enlarge it 
by 5\% on each side to simulate slight localization uncertainty.

\subsection{Datasets and Models}
\label{sec:datasets}

\noindent\textbf{Datasets.} We use three polyp segmentation benchmarks: 
CVC-300~\cite{vazquez2017cvc300} (60 images), 
CVC-ColonDB~\cite{tajbakhsh2015colondb} (380 images), and 
ETIS-LaribPolypDB~\cite{silva2014etis} (196 images). All three are 2D RGB 
colonoscopy datasets; generalizability to other modalities or anatomies is 
addressed in Sec.~\ref{sec:conclusion}.

\begin{table}[htbp]
  \caption{Summary of polyp segmentation datasets used in our benchmark.}
  \label{tab:datasets}
  \centering
  \begin{tabular}{@{}lccc@{}}
    \toprule
    Dataset & Modality & Object & Masks \\
    \midrule
    CVC-300~\cite{vazquez2017cvc300} & Endoscopy & Polyp & 60 \\
    CVC-ColonDB~\cite{tajbakhsh2015colondb} & Endoscopy & Polyp & 380 \\
    ETIS-LaribPolypDB~\cite{silva2014etis} & Endoscopy & Polyp & 196 \\
    \bottomrule
  \end{tabular}
\end{table}

\noindent\textbf{Models.} We benchmark seven models in two groups: 
\emph{General-purpose}: SAM~\cite{kirillov2023sam} (ViT-H), 
SAM~2~\cite{ravi2024sam2} (Hiera-L), SAM~3~\cite{carion2025sam3segmentconcepts}. 
\emph{Medical-adapted}: MedSAM~\cite{ma2024medsam}, 
SAM-Med2D~\cite{cheng2023sammed2d}, MedSAM2~\cite{zhu2024medsam2}, 
MedSAM3~\cite{liu2025medsam3delvingsegmentmedical}. To the best of our 
knowledge, none were explicitly fine-tuned on our target datasets. We use 
only publicly available pretrained weights with no additional training or 
dataset-specific tuning.


\begin{table*}[htbp]
  \caption{Summary of evaluated models. All models are assessed in a 
  zero-shot manner, i.e., without fine-tuning or adaptation on the target 
  endoscopy datasets.}
  \label{tab:models}
  \centering
  \small
  \resizebox{\textwidth}{!}{
  \begin{tabular}{@{}llcccc@{}}
    \toprule
    Setting & Model & Year & Backbone / Initialization & Training Scale & Data Type \\
    \midrule
    \multirow{3}{*}{General-purpose}
    & SAM~\cite{kirillov2023sam} & 2023 & ViT-H & 11M images, 1B masks & Natural 2D images \\
    & SAM~2~\cite{ravi2024sam2} & 2024 & Hiera-L & 50.9K videos, 642K masklets & Natural images + videos \\
    & SAM~3~\cite{carion2025sam3segmentconcepts} & 2026 & VLP-based encoder & 5.2M HQ images, 52.5K videos & Natural images + videos (open-vocab) \\
    \midrule
    \multirow{4}{*}{Medical-adapted}
    & MedSAM~\cite{ma2024medsam} & 2023 & SAM (ViT-H) fine-tuned & 1.6M medical images & Medical 2D images \\
    & SAM-Med2D~\cite{cheng2023sammed2d} & 2023 & SAM (ViT-H) + PEFT & 4.6M images, 19.7M masks & Medical 2D images \\
    & MedSAM2~\cite{zhu2024medsam2} & 2024 & SAM2-Tiny fine-tuned & 455K 3D pairs, 76K frames & Medical 3D images + video \\
    & MedSAM3~\cite{liu2025medsam3delvingsegmentmedical} & 2025 & SAM3-based fine-tuned & 658K images, 2.86M instances & Medical 2D \& 3D images \\
    \bottomrule
  \end{tabular}
  }
\end{table*}
\section{Experimental Results}
\label{sec:results}

All models are evaluated in their released configurations without fine-tuning. The factorial design spans $2$ occlusion types $\times$ $4$ severity levels $\times$ $3$ mask targets $\times$ $2$ prompt types $\times$ $7$ models $\times$ $3$ datasets $=1{,}008$ configurations. Relative degradation is computed as $\Delta(\%) = (\text{DSC}_{\text{clean}} - \text{DSC}_{\text{occ}})/ \text{DSC}_\text{clean} \times 100$. Unless stated otherwise, we report box prompt results under surgical tool occlusion as the primary condition, as it is the most clinically motivated and reveals the sharpest behavioral differences. HD95 results are included in supplementary material.

We evaluate all seven models across three datasets, two occlusion types, 
three severity levels, and two prompt types. Unless stated otherwise, we 
report box-prompt results under surgical tool occlusion as the primary 
condition, as it is the most clinically motivated and reveals the sharpest 
behavioral differences. All results are reported in DSC; HD95 results are 
included in supplementary material.

\subsection{Full-Mask Evaluation}
\label{sec:full_analysis}

\begin{table*}[t]
\centering
\caption{\textbf{Full (amodal) DSC under surgical tool occlusion, box prompts.} MedSAM2 appears the most robust model under this metric. However, as shown in Sec.~\ref{sec:visible_analysis}, this ranking is an artifact: high full-mask scores can result from models that segment through the occluder rather than around it. $\Delta$\% (lower is better) is relative degradation from clean to high severity. \best{Best} and \second{second best} highlighted per column per dataset.}
\label{tab:full_mask}
\small
\setlength{\tabcolsep}{3.5pt}
\begin{tabular}{l|ccccc|ccccc|ccccc}
\toprule
& \multicolumn{5}{c|}{\textbf{CVC-300}} 
& \multicolumn{5}{c|}{\textbf{CVC-ColonDB}} 
& \multicolumn{5}{c}{\textbf{ETIS}} \\
\textbf{Model} 
& Clean & Low & Med & High & $\Delta$\% 
& Clean & Low & Med & High & $\Delta$\% 
& Clean & Low & Med & High & $\Delta$\% \\
\midrule
SAM
& .925 & .849 & .754 & .583 & 37.0
& .884 & .800 & .599 & .567 & 35.9
& .910 & .807 & .662 & .552 & 39.4 \\
SAM~2
& .913 & .865 & .766 & \second{.692} & 24.2
& .907 & .844 & .731 & .651 & 28.2
& .907 & .838 & .736 & .621 & 31.6 \\
SAM~3
& \best{.950} & \second{.869} & .776 & .618 & 35.0
& \best{.928} & \best{.868} & .762 & .631 & 32.0
& \best{.935} & .845 & \second{.762} & .603 & 35.6 \\
\midrule
MedSAM
& .742 & .635 & .602 & .640 & \best{13.8}
& .709 & .664 & .573 & .607 & \best{14.5}
& .778 & .724 & .622 & .588 & \second{24.5} \\
SAM-Med2D
& .903 & .845 & .699 & .554 & 38.7
& .854 & .762 & .568 & .545 & 36.2
& .837 & .780 & .610 & .502 & 40.0 \\
MedSAM2
& .930 & \best{.911} & \best{.833} & \best{.784} & \second{15.7}
& \second{.912} & \best{.868} & \best{.818} & \best{.758} & \second{17.0}
& .907 & \best{.874} & \best{.810} & \best{.766} & \best{15.6} \\
MedSAM3
& \second{.937} & .845 & \second{.793} & .675 & 27.9
& .908 & \second{.855} & \second{.766} & \second{.659} & 27.4
& \second{.918} & \second{.867} & .760 & \second{.680} & 26.0 \\
\bottomrule
\end{tabular}
\end{table*}

Table~\ref{tab:full_mask} presents the standard full-mask evaluation. At 
first glance, MedSAM2 appears to be the strongest model at all occlusion 
levels, exhibiting the smallest degradation ($\sim$15--17\%) across all 
three datasets. MedSAM shows a counterintuitive pattern: its full-mask DSC 
\emph{increases} from medium to high occlusion on CVC-300 
(0.602$\rightarrow$0.640), which appears to indicate robustness. SAM~3, 
despite achieving the best clean performance on all datasets, degrades more 
substantially under high occlusion, falling behind not only MedSAM2 but 
also SAM~2 on CVC-300 (0.618 vs.\ 0.692). These observations are artifacts 
of the full-mask metric, as the rankings reverse substantially once 
visible-only DSC is examined.

\subsection{Visible-Only Evaluation}
\label{sec:visible_analysis}

\begin{table*}[t]
\centering
\caption{\textbf{Visible-only DSC under surgical tool occlusion, box 
prompts.} SAM~3 achieves the highest visible-region DSC in the majority of columns, with MedSAM2 and MedSAM3 competitive across datasets. MedSAM and SAM-Med2D show the steepest degradation under increasing occlusion severity. $\Delta$\% (lower is better) is relative degradation from clean to high severity. \best{Best} and \second{second best} highlighted per column per dataset.}
\label{tab:visible_mask}
\small
\setlength{\tabcolsep}{3.5pt}
\begin{tabular}{l|ccccc|ccccc|ccccc}
\toprule
& \multicolumn{5}{c|}{\textbf{CVC-300}} & \multicolumn{5}{c|}{\textbf{CVC-ColonDB}} & \multicolumn{5}{c}{\textbf{ETIS}} \\
\textbf{Model} & Clean & Low & Med & High & $\Delta$\% & Clean & Low & Med & High & $\Delta$\% & Clean & Low & Med & High & $\Delta$\% \\
\midrule
SAM      
& .925 & .886 & \second{.914} & .388 & 58.1 
& .884 & .844 & .635 & .505 & 42.9 
& .910 & .842 & .783 & .554 & 39.1 \\
SAM~2     
& .913 & .910 & .894 & .662 & 27.5 
& .907 & \second{.887} & .824 & \second{.730} & \second{19.5} 
& .907 & .877 & \second{.863} & .725 & 20.1 \\
SAM~3     
& \best{.950} & \second{.913} & \best{.931} & \best{.724} & \best{23.8} 
& \best{.928} & \best{.912} & \best{.876} & \best{.748} & \best{19.3} 
& \best{.935} & .890 & \best{.887} & \best{.801} & \best{14.3} \\
\midrule
MedSAM   
& .742 & .669 & .579 & .302 & 59.3 
& .709 & .659 & .462 & .281 & 60.4 
& .778 & .759 & .598 & .395 & 49.2 \\
SAM-Med2D 
& .903 & .889 & .780 & .345 & 61.8 
& .854 & .801 & .544 & .385 & 54.9 
& .837 & .818 & .654 & .491 & 41.3 \\
MedSAM2   
& .930 & \best{.937} & .870 & \second{.687} & \second{26.1} 
& \second{.912} & .873 & .750 & .586 & 35.7 
& .907 & \second{.893} & .797 & .638 & 29.7 \\
MedSAM3   
& \second{.937} & .882 & .893 & .661 & 29.4 
& .908 & .885 & \second{.836} & .723 & 20.3 
& \second{.918} & \best{.914} & .841 & \second{.743} & \second{19.1} \\
\bottomrule
\end{tabular}
\end{table*}
Table~\ref{tab:visible_mask} paints a very different picture. The full-mask ranking collapses: MedSAM's apparently competitive score of 0.640 at high occlusion on CVC-300 is revealed to be misleading; its visible DSC is only 0.302, meaning most of its ``correct'' predictions fell inside the tool footprint, not on actual tissue. The key patterns are:

\noindent\textbf{SAM~3 is the most robust model across all datasets.} 
SAM~3 holds the top position at high severity on all three datasets (0.724, 
0.748, 0.801) and achieves the lowest degradation overall. MedSAM2 emerges 
as a strong second on CVC-300 (0.687 vs.\ SAM~3's 0.724), while SAM~2 
takes second on CVC-ColonDB and ETIS (0.730 and 0.725 respectively). 
MedSAM3 remains competitive but does not consistently occupy the second 
position across datasets, finishing third on CVC-300 (0.661) behind both 
SAM~3 and MedSAM2. SAM degrades sharply at high severity on CVC-300 
(0.388), despite a surprisingly strong medium-severity score (0.914, second 
best overall on that column).

\noindent\textbf{{Domain-adapted models suffer catastrophic degradation.}} Medically adapted architectures built on SAM 1/2 (MedSAM, SAM-Med2D) exhibit severe performance collapse, with visible DSC degrading by 41–60\% under high occlusion. This suggests a critical vulnerability introduced during the domain adaptation process. Because these models are fine-tuned almost exclusively on highly curated, clean medical datasets, they likely overfit to pristine, unobstructed tissue boundaries. Consequently, the introduction of high-frequency surgical tool edges acts as an out-of-distribution adversarial perturbation, causing the models to either aggressively leak into the tool region or catastrophically fail to segment the visible tissue entirely. This exposes a major blind spot in current medical foundation model training paradigms: optimizing for clean-image performance severely degrades compositional robustness.

\noindent\textbf{MedSAM3 partially preserves SAM~3's robustness, but with inconsistent second-place finishes.} Unlike MedSAM and SAM-Med2D, LoRA-based adaptation of SAM~3 does not collapse robustness, and MedSAM3 remains far above MedSAM and SAM-Med2D across all datasets. However, it is outperformed at high severity by MedSAM2 on CVC-300 and by SAM~2 on CVC-ColonDB and ETIS. One possible explanation is that SAM~3's architectural prior, rather than medical fine-tuning itself, is the primary driver of robustness.

\noindent\textbf{SAM-Med2D underperforms across all conditions.} 
SAM-Med2D achieves neither the robustness of SAM~3/MedSAM3 nor meaningful 
amodal tendency, consistently placed below both archetypes across all 
datasets and severity levels.

\subsection{Invisible-Region Evaluation}
\label{sec:invisible_analysis}

\begin{table}[t]
\centering
\caption{\textbf{Invisible-region DSC under surgical tool occlusion (box prompts).}
Higher values indicate prediction inside the occluded region. At high severity, scores may be inflated by the large tool footprint and should be interpreted together with visible DSC (Table~\ref{tab:visible_mask}). \best{Best} and \second{second best} are highlighted per column.}

\label{tab:invisible_mask}
\small
\renewcommand{\arraystretch}{1.3}
\resizebox{\columnwidth}{!}{
\setlength{\tabcolsep}{4pt}
\begin{tabular}{l|ccc|ccc|ccc}
\toprule
& \multicolumn{3}{c|}{\textbf{CVC-300}} & \multicolumn{3}{c|}{\textbf{CVC-ColonDB}} & \multicolumn{3}{c}{\textbf{ETIS}} \\
\textbf{Model} & Low & Med & High & Low & Med & High & Low & Med & High \\
\midrule
SAM       
& .039 & .002 & .588 
& .052 & .285 & .501 
& .035 & .155 & .442 \\
SAM~2     
& .120 & .190 & .569 
& .114 & .278 & .410 
& .122 & .158 & .395 \\
SAM~3     
& .048 & .056 & .336 
& .082 & .168 & .304 
& .035 & .106 & .185 \\
\midrule
MedSAM    
& \second{.387} & \second{.587} & \best{.823} 
& \best{.492} & \second{.643} & \best{.787} 
& \second{.222} & \second{.546} & \second{.695} \\
SAM-Med2D 
& .065 & .297 & .657 
& .133 & .419 & .629 
& .098 & .304 & .478 \\
MedSAM2   
& \best{.487} & \best{.657} & \second{.816} 
& \second{.478} & \best{.724} & \second{.736} 
& \best{.358} & \best{.562} & \best{.707} \\
MedSAM3   
& .118 & .214 & .452 
& .223 & .295 & .406 
& .155 & .272 & .440 \\
\bottomrule
\end{tabular}
}
\end{table}

Table~\ref{tab:invisible_mask} shows prediction overlap with the hidden 
polyp region beneath the tool. Interpreting these scores requires care, 
particularly as severity increases.

\noindent\textbf{At low and medium severity, this is a more reliable 
signal.} When the tool footprint is small, invisible DSC provides a clearer signal of amodal tendency. MedSAM2 consistently achieves the highest scores across datasets, with MedSAM close behind, suggesting these medical-domain models more frequently predict into occluded regions. In contrast, SAM~3 produces near-zero scores in several cases (e.g., 0.056 on CVC-300 at medium severity), indicating a stronger tendency to suppress predictions when the target becomes partially hidden.

\noindent\textbf{High severity: geometric coincidence confounder.} At High occlusion levels, the instrument footprint constitutes a massive proportion of the target area. Under these conditions, the invisible DSC ($M_{inv}$) becomes highly susceptible to geometric coincidence: any unconstrained, over-segmented prediction that bleeds into the tool region will artificially inflate the amodal score. For instance, MedSAM attains high invisible scores despite suffering poor visible DSC, indicating that its "amodal completion" is merely uncontrolled boundary overflow rather than deliberate structural reasoning. Therefore, we establish a critical heuristic for this benchmark: high invisible DSC can only be interpreted as true amodal completion if the model simultaneously maintains a high visible DSC. Under this strict criterion, MedSAM2 is the only evaluated architecture that demonstrates genuine, boundary-consistent amodal behavior under severe occlusion. 

\begin{figure*}[b]
  \centering
  \includegraphics[width=0.85\linewidth]{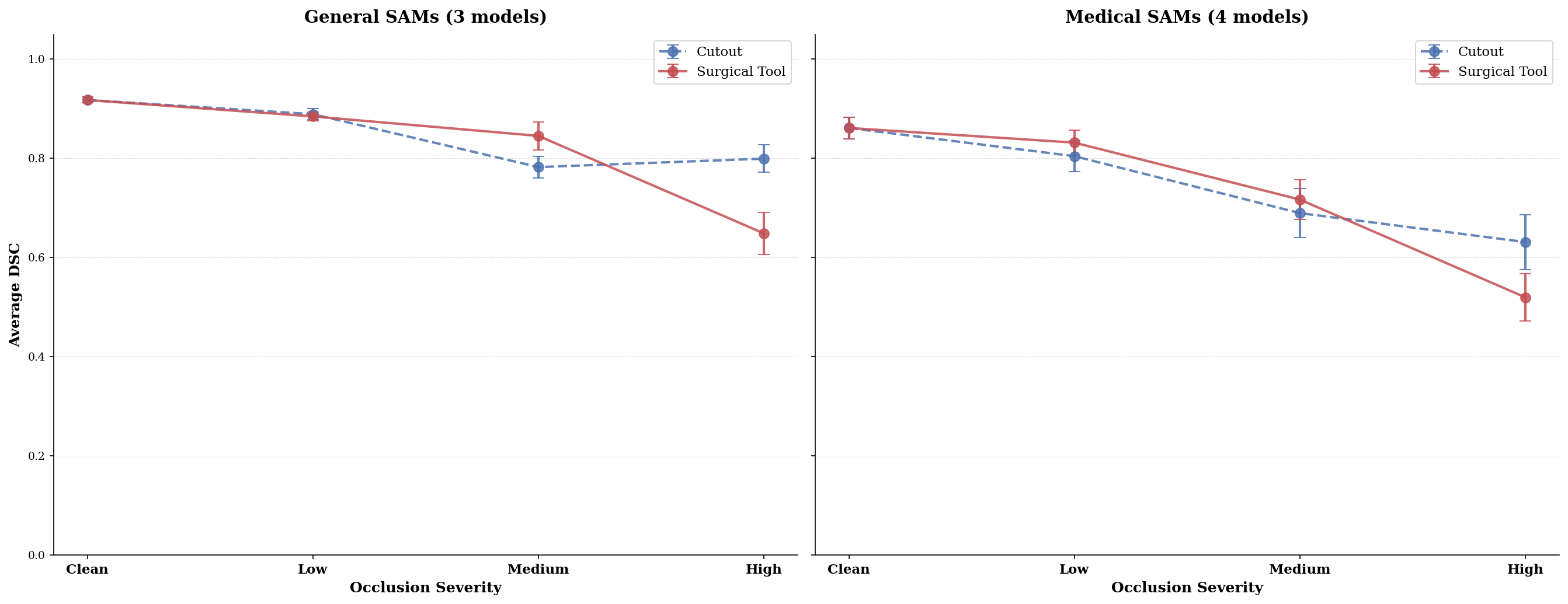}
  \caption{Visible-only DSC comparison between Cutout and Surgical Tool 
  occlusions across 3 datasets under increasing severity (Clean, Low, 
  Medium, High) with Box prompts. \textit{Left}: General SAMs (SAM, SAM~2, SAM~3); \textit{Right}: Medical SAMs (MedSAM, SAM-Med2D, MedSAM2, MedSAM3). Differences between occlusion types are minimal at low-to-moderate levels but diverge at High severity, where surgical occlusion induces substantially larger degradation.}
  \label{fig:occlusion_type_comparison}
\end{figure*}

\noindent\textbf{Behavioral archetypes.}
Taken together, Tables~\ref{tab:visible_mask} and~\ref{tab:invisible_mask} suggest that models tend to exhibit two behaviors:

\begin{itemize}[leftmargin=*,itemsep=2pt]
    \item \textbf{Occluder-Aware:} SAM, SAM~2, SAM~3, and MedSAM3 largely suppress predictions inside the tool region, yielding higher visible DSC. These models are preferable when the goal is accurate delineation of \emph{visible} anatomy.
    
    \item \textbf{Occluder-Agnostic:} MedSAM and MedSAM2 frequently predict into occluded regions, reflected by consistently higher invisible DSC at low and medium severity. This pattern suggests a tendency toward amodal completion, though at high severity the signal becomes confounded by instrument overlap.
    
    \item \textbf{Underperforming:} SAM-Med2D performs poorly on both
    metrics and does not clearly exhibit either behavior.
\end{itemize}

\noindent\textbf{MedSAM2 as a boundary case.}
Within the Occluder-Agnostic group, MedSAM2 stands out by combining strong invisible-region scores with competitive visible DSC. This balanced profile suggests predictions that extend into occluded regions while remaining more boundary-consistent than MedSAM, potentially reflecting its video-based fine-tuning strategy~\cite{zhu2024medsam2}. Such behavior may be useful in applications where both visible accuracy and estimation of hidden anatomy are relevant.

\subsection{Effect of Occlusion Types}
\label{sec:occlusion_type}


\noindent\textbf{Segmentation performance decreases as occlusion 
severity increases, regardless of occlusion type.} As shown in 
Fig.~\ref{fig:occlusion_type_comparison}, both Cutout and Surgical 
Tool occlusions exhibit a consistent downward trend from Clean to High 
severity across General and Medical SAMs, starting from a strong 
baseline ($0.918$ and $0.861$ DSC for General and Medical groups 
respectively), indicating that severity level is the primary driver 
of performance degradation.

\noindent\textbf{Occlusion type has minimal impact at low severity 
but diverges substantially at high severity.} The two occlusion types 
produce nearly identical results at Clean and Low severity. At Medium 
severity, Surgical Tool marginally outperforms Cutout for both model 
groups. At High severity, however, this trend reverses: Surgical Tool 
induces substantially larger degradation than Cutout ($0.649$ vs. 
$0.799$ for General SAMs; $0.520$ vs. $0.631$ for Medical SAMs), 
indicating that occlusion type interacts with severity rather than 
contributing a constant offset.

\subsection{Box \textit{vs.}\ Single-Point Prompt Sensitivity}
\label{sec:prompt_analysis}

\begin{figure*}[t]
  \centering
  \includegraphics[width=0.9\linewidth]{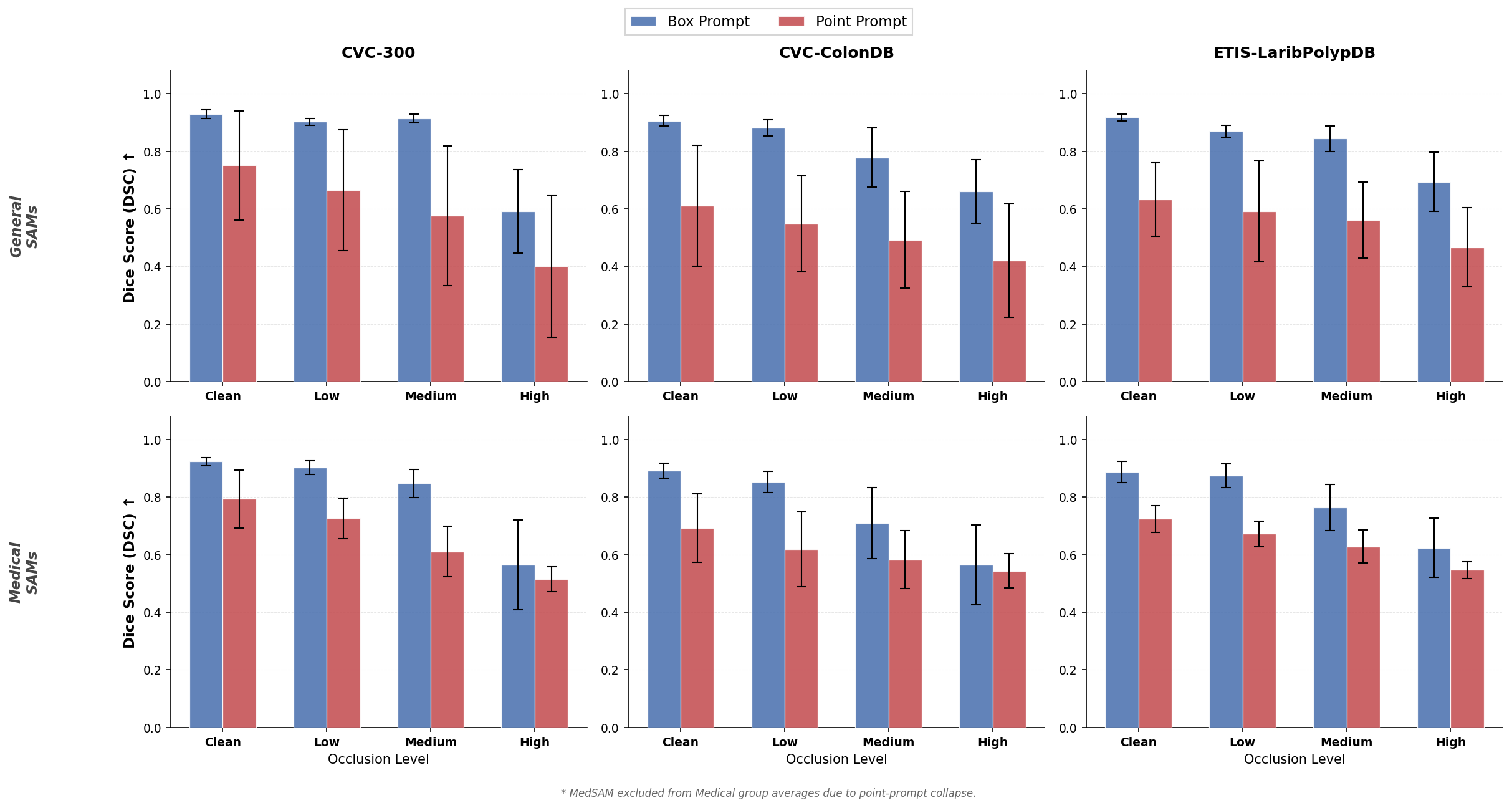}
  \caption{Visible-mask DSC comparison between Box and Single-Point prompts 
  under increasing occlusion levels (Clean, Low, Medium, High), with 
  surgical tool occlusions. \textit{Top row}: General SAMs (SAM, SAM~2, 
  SAM~3); \textit{bottom row}: Medical SAMs (SAM-Med2D, MedSAM2, 
  MedSAM3). Columns correspond to CVC-300, CVC-ColonDB, and 
  ETIS-LaribPolypDB. Box prompting demonstrates consistently higher DSC 
  and more consistent performance across datasets under all occlusion 
  levels.}
  \label{fig:box_vs_point}
\end{figure*}

\begin{table}[htbp]
\caption{\textbf{Visible-only DSC under box vs. point prompts.}
Clean Avg is performance on clean images; Occluded Avg averages
Low, Medium, and High occlusion severities across both occlusion types.
MedSAM collapses under point prompting but remains functional with box prompts.}
\label{tab:medsam_collapse}
\centering
\small
\setlength{\tabcolsep}{6pt}
\begin{tabular}{lccc}
\toprule
\textbf{Model} & \textbf{Prompt} & \textbf{Clean Avg} & \textbf{Occluded Avg} \\
\midrule
MedSAM   & Box   & 0.7431 & 0.4872 \\
MedSAM   & Point & 0.0207 & 0.0172 \\
\midrule
MedSAM3  & Point & 0.8429 & 0.6840 \\
SAM 3    & Point & 0.8217 & 0.7046 \\
\bottomrule
\end{tabular}
\end{table}

\noindent\textbf{MedSAM collapses under single-point prompting.}
Unlike other models, MedSAM degrades to near-zero DSC with a single interior point 
(Clean Avg: 0.02; Occluded Avg: 0.017) but remains functional with box prompts 
(Occluded Avg: 0.487). This aligns with its training protocol, which fine-tunes the 
prompt encoder only on bounding box inputs~\cite{ma2024medsam}. Point prompts are 
therefore out-of-distribution and cause systematic failure rather than general 
performance degradation. Consistent with prior work~\cite{chang2025lumbar}, we report 
MedSAM results using box prompts only.

\noindent\textbf{Box prompting consistently outperforms point prompting.}
As shown in Fig.~\ref{fig:box_vs_point}, box prompts achieve higher DSC across all 
datasets and occlusion levels (averaged over six models excluding MedSAM). The gap 
is larger for general SAM models than for medical variants, suggesting domain 
fine-tuning reduces but does not eliminate prompt sensitivity. Under severe 
occlusion, the advantage of box prompts narrows for medical models (especially 
SAM-Med2D), indicating reduced spatial grounding benefits. Among point prompts, 
MedSAM3 achieves the best DSC ahead of SAM~3, while SAM~3 
performs best under occlusion, both outperforming 
MedSAM2. This reversal suggests SAM~3's architectural robustness 
becomes more prominent under occlusion.

\section{Conclusion and Discussion}
\label{sec:conclusion}

We introduced \textbf{OccSAM-Bench}, the first targeted benchmark designed to systematically evaluate the robustness of SAM-family foundation models under controlled surgical occlusion in endoscopic environments. Our findings demonstrate that the reliance on standard full-mask Dice scores in medical segmentation is not only misleading but potentially hazardous: it routinely rewards models that erroneously hallucinate tissue through surgical instruments while penalizing those that correctly suppress predictions.

By deploying our novel three-region evaluation protocol, we exposed a critical divergence in how modern architectures process partial visibility. We identified two distinct behavioral archetypes: \textit{Occluder-Aware} models (SAM, SAM 2, SAM 3, MedSAM3), which prioritize the conservative, safe delineation of visible tissue; and \textit{Occluder-Agnostic} models (MedSAM, MedSAM2), which exhibit amodal-like completion tendencies by predicting into occluded territories. Notably, MedSAM2 strikes a unique balance, retaining competitive visible-tissue accuracy while predicting hidden structures. Ultimately, these archetypes dictate that model selection must be treated as a clinical design choice, prioritizing conservative boundaries for safe robotic navigation versus amodal inference for hidden anatomical estimation.

\textbf{Limitations and Future Work.} We explicitly acknowledge that OccSAM-Bench relies on 2D RGB colonoscopy data and simplified synthetic occlusions (cutouts and 2D tool overlays). While this synthetic approach was methodologically necessary to achieve precise mathematical control over occlusion severity against a known ground truth, it inherently bypasses the complex optical physics of true in-vivo occlusion, such as dynamic lighting alterations, specular highlights, and instrument-induced tissue deformation. Extending this evaluation paradigm to encompassing diverse modalities, physically realistic rendering, and video-based temporal tracking across complex surgical interventions remains a vital trajectory for future research.

\section{Acknowledgment}
This work has been partially supported by the NIH grants:
R01-HL171376 and U01-CA268808.\\

{
    \small
    \bibliographystyle{ieeenat_fullname}
    \bibliography{main}
}


\clearpage
\appendix
\setcounter{page}{1}
\onecolumn
\raggedbottom  
\section*{Supplementary Material}
\setlength{\tabcolsep}{3pt}
\renewcommand{\arraystretch}{1.05}

We report additional quantitative results across three datasets under 
different occlusion generation strategies and prompt settings. 
Tables~\ref{tab:sup_cvc300_tool_box}--\ref{tab:sup_cvc300_cutout_point} 
present results on CVC-300, Tables~\ref{tab:sup_colondb_tool_box}-\ref{tab:sup_colondb_cutout_point} on CVC-ColonDB, and Tables~\ref{tab:sup_etis_tool_box}--\ref{tab:sup_etis_cutout_point} on ETIS-LaribPolypDB. Each table reports Dice Similarity Coefficient (DSC, $\uparrow$) and 95th percentile Hausdorff Distance (HD95, $\downarrow$) 
under four severity levels (Clean, Low, Medium, High), evaluated across 
three modes: \textit{Full (Amodal)}, \textit{Visible Only}, and 
\textit{Invisible (Occluded)}. Results consistently show that performance degrades with increasing occlusion severity, with the Invisible mode being the most challenging as models must infer occluded regions without direct visual evidence. Box prompts generally yield stronger results than Point prompts across all settings.

\begin{table}[H]
\centering
\footnotesize
\caption{Segmentation performance on CVC-300 under Surgical Tool occlusion using Box prompts across clean, low, medium, and high severity levels. \textbf{Bold} and \underline{underline} denote the best and second best per column, respectively.}
\label{tab:sup_cvc300_tool_box}
\begin{tabular}{llcccccccc}
\toprule
\multirow{2}{*}{Eval Mode} & \multirow{2}{*}{Model} 
& \multicolumn{2}{c}{Clean} 
& \multicolumn{2}{c}{Low} 
& \multicolumn{2}{c}{Medium} 
& \multicolumn{2}{c}{High} \\
\cmidrule(lr){3-4} \cmidrule(lr){5-6} \cmidrule(lr){7-8} \cmidrule(lr){9-10}
& & DSC $\uparrow$ & HD95 $\downarrow$ 
  & DSC $\uparrow$ & HD95 $\downarrow$ 
  & DSC $\uparrow$ & HD95 $\downarrow$ 
  & DSC $\uparrow$ & HD95 $\downarrow$ \\
\midrule
\multirow{7}{*}{Full (Amodal)}
& SAM        & 0.93 & 10.05 & 0.85 & 38.30 & 0.75 & 62.17 & 0.58 & 73.95 \\
& SAM~2      & 0.91 & 8.48  & \second{0.87} & \second{35.34} & 0.77 & 61.64 & \second{0.69} & 65.53 \\
& SAM~3      & \best{0.95} & \best{4.98} & 0.87 & 35.51 & 0.78 & 62.42 & 0.62 & \second{60.11}  \\
& MedSAM     & 0.74 & 24.43 & 0.63 & 53.34 & 0.60 & 68.74 & 0.64 & 64.27 \\
& SAM-Med2D  & 0.90 & 12.78 & 0.84 & 36.47 & 0.70 & 66.66 & 0.55 & 75.47 \\
& MedSAM2    & 0.93 & 7.40  & \best{0.91} & \best{22.85} & \best{0.83} & \best{42.28} & \best{0.78} & \best{51.21}  \\
& MedSAM3    & \second{0.94} & \second{6.69} & 0.85 & 38.06 & \second{0.79} & \second{61.30} & 0.68 & 62.09 \\
\midrule
\multirow{7}{*}{Visible Only}
& SAM        & 0.93 & 10.05 & 0.89 & 22.67 & \second{0.91} & 17.72 & 0.39 & 61.41 \\
& SAM~2      & 0.91 & 8.48  & \second{0.91} & \second{16.27} & 0.89 & \best{10.92} & 0.66 & \second{31.65}  \\
& SAM~3      & \best{0.95} & \best{4.98} & \second{0.91} & 16.40 & \best{0.93} & \second{12.80} & \best{0.72} & \best{27.53}  \\
& MedSAM     & 0.74 & 24.43 & 0.67 & 48.50 & 0.58 & 54.28 & 0.30 & 80.01 \\
& SAM-Med2D  & 0.90 & 12.78 & 0.89 & 23.57 & 0.78 & 34.57 & 0.35 & 71.11 \\
& MedSAM2    & 0.93 & 7.40  & \best{0.94} & \best{15.41} & 0.87 & 22.11 & \second{0.69} & 33.96 \\
& MedSAM3    & \second{0.94} & \second{6.69} & 0.88 & 23.72 & 0.89 & 24.84 & 0.66 & 40.74 \\
\midrule
\multirow{7}{*}{Invisible (Occluded)}
& SAM        & -- & -- & 0.04 & 62.98  & 0.00 & 105.32 & 0.59 & 47.30 \\
& SAM~2      & -- & -- & 0.12 & 43.98  & 0.19 & 61.38  & 0.57 & 50.49 \\
& SAM~3      & -- & -- & 0.05 & 68.19  & 0.06 & 71.62  & 0.34 & 65.63 \\
& MedSAM     & -- & -- & \second{0.39} & \second{32.57} & \second{0.59} & \best{34.51} & \best{0.82} & \best{29.25}  \\
& SAM-Med2D  & -- & -- & 0.07 & 53.41  & 0.30 & 67.90  & \second{0.66} & 48.26 \\
& MedSAM2    & -- & -- & \best{0.49} & \best{25.90} & \best{0.66} & \second{34.80} & \best{0.82} & \second{37.88}  \\
& MedSAM3    & -- & -- & 0.12 & 33.08  & 0.21 & 48.52  & 0.45 & 54.07 \\
\bottomrule
\end{tabular}
\end{table}

\begin{table}[H]
\centering
\footnotesize
\caption{Segmentation performance on CVC-300 under Cutout occlusion using Box prompts across clean, low, medium, and high severity levels. \textbf{Bold} and \underline{underline} denote the best and second best per column, respectively.}
\label{tab:sup_cvc300_cutout_box}
\begin{tabular}{llcccccccc}
\toprule
\multirow{2}{*}{Eval Mode} & \multirow{2}{*}{Model} 
& \multicolumn{2}{c}{Clean} 
& \multicolumn{2}{c}{Low} 
& \multicolumn{2}{c}{Medium} 
& \multicolumn{2}{c}{High} \\
\cmidrule(lr){3-4} \cmidrule(lr){5-6} \cmidrule(lr){7-8} \cmidrule(lr){9-10}
& & DSC $\uparrow$ & HD95 $\downarrow$ 
  & DSC $\uparrow$ & HD95 $\downarrow$ 
  & DSC $\uparrow$ & HD95 $\downarrow$ 
  & DSC $\uparrow$ & HD95 $\downarrow$ \\
\midrule
\multirow{7}{*}{Full (Amodal)}
& SAM        & 0.93 & 10.05 & 0.89 & 17.48 & 0.69 & 35.73 & 0.61 & 36.59 \\
& SAM~2      & 0.91 & 8.48  & 0.92 & 15.81 & 0.77 & 31.93 & 0.68 & 31.15 \\
& SAM~3      & \best{0.95} & \best{4.98} & \best{0.94} & \best{6.95} & \best{0.89} & \second{19.67} & \best{0.87} & 27.55 \\
& MedSAM     & 0.74 & 24.43 & 0.68 & 30.31 & 0.60 & 37.29 & 0.64 & 32.04 \\
& SAM-Med2D  & 0.90 & 12.78 & 0.87 & 20.81 & 0.75 & 30.07 & 0.68 & 32.14 \\
& MedSAM2    & 0.93 & 7.40  & 0.91 & 14.18 & 0.82 & 24.12 & 0.77 & \second{25.65}  \\
& MedSAM3    & \second{0.94} & \second{6.69} & \second{0.93} & \second{9.47} & \second{0.86} & \best{16.13} & \second{0.84} & \best{23.00}  \\
\midrule
\multirow{7}{*}{Visible Only}
& SAM        & 0.93 & 10.05 & \best{0.92} & 13.11 & 0.74 & 19.96 & 0.78 & 20.05 \\
& SAM~2      & 0.91 & 8.48  & \best{0.92} & \best{8.93} & \second{0.84} & \best{16.34} & \best{0.92} & \best{6.70}  \\
& SAM~3      & \best{0.95} & \best{4.98} & \second{0.91} & 14.09 & 0.81 & 23.55 & \second{0.79} & 25.63 \\
& MedSAM     & 0.74 & 24.43 & 0.63 & 25.56 & 0.35 & 33.63 & 0.25 & 34.01 \\
& SAM-Med2D  & 0.90 & 12.78 & 0.86 & 15.83 & 0.75 & 19.91 & 0.75 & 20.51 \\
& MedSAM2    & 0.93 & 7.40  & \second{0.91} & \second{10.57} & \best{0.85} & \second{17.13} & \second{0.79} & \second{19.85}  \\
& MedSAM3    & \second{0.94} & \second{6.69} & 0.90 & 18.41 & 0.75 & 31.61 & 0.75 & 28.30 \\
\midrule
\multirow{7}{*}{Invisible (Occluded)}
& SAM        & -- & -- & 0.15 & 17.77 & 0.24 & 27.74 & 0.24 & 20.21 \\
& SAM~2      & -- & -- & 0.54 & 10.44 & 0.24 & 25.36 & 0.21 & 44.04 \\
& SAM~3      & -- & -- & 0.77 & 6.92  & 0.78 & \second{10.29} & \second{0.78} & \second{13.45}  \\
& MedSAM     & -- & -- & \second{0.82} & \best{3.58} & \best{0.90} & \best{8.60} & \best{0.86} & \best{10.46}  \\
& SAM-Med2D  & -- & -- & 0.39 & 10.51 & 0.46 & 16.71 & 0.45 & 26.40 \\
& MedSAM2    & -- & -- & 0.50 & 7.74  & 0.49 & 16.66 & 0.51 & 30.44 \\
& MedSAM3    & -- & -- & \best{0.88} & \second{6.00} & \second{0.80} & 11.89 & 0.72 & 19.87 \\
\bottomrule
\end{tabular}
\end{table}

\begin{table}[H]
\centering
\footnotesize
\caption{Segmentation performance on CVC-300 under Surgical Tool occlusion using Point prompts across clean, low, medium, and high severity levels. \textbf{Bold} and \underline{underline} denote the best and second best per column, respectively.}
\label{tab:sup_cvc300_tool_point}
\begin{tabular}{llcccccccc}
\toprule
\multirow{2}{*}{Eval Mode} & \multirow{2}{*}{Model} 
& \multicolumn{2}{c}{Clean} 
& \multicolumn{2}{c}{Low} 
& \multicolumn{2}{c}{Medium} 
& \multicolumn{2}{c}{High} \\
\cmidrule(lr){3-4} \cmidrule(lr){5-6} \cmidrule(lr){7-8} \cmidrule(lr){9-10}
& & DSC $\uparrow$ & HD95 $\downarrow$ 
  & DSC $\uparrow$ & HD95 $\downarrow$ 
  & DSC $\uparrow$ & HD95 $\downarrow$ 
  & DSC $\uparrow$ & HD95 $\downarrow$ \\
\midrule
\multirow{7}{*}{Full (Amodal)}
& SAM        & 0.49 & 165.54 & 0.23 & 468.78 & 0.18 & 517.32 & 0.39 & 206.37 \\
& SAM~2      & \second{0.83} & 44.03  & 0.66 & 130.22 & 0.59 & 144.94 & 0.49 & 130.24 \\
& SAM~3      & \best{0.93} & \best{6.71} & \best{0.73} & 105.00 & 0.52 & 112.44 & 0.50 & 129.06 \\
& MedSAM     & 0.02 & 89.81  & 0.00 & 180.03 & 0.01 & 177.41 & 0.02 & 150.37 \\
& SAM-Med2D  & 0.68 & 40.67  & 0.57 & \best{79.31} & 0.38 & \second{99.08} & \best{0.51} & \best{99.78}  \\
& MedSAM2    & 0.77 & 25.96  & 0.66 & \second{83.86} & \second{0.60} & \best{86.40} & \best{0.51} & \second{111.29}  \\
& MedSAM3    & \best{0.93} & \second{8.24} & \second{0.73} & 120.35 & \best{0.68} & 103.49 & \second{0.49} & 120.83 \\
\midrule
\multirow{7}{*}{Visible Only}
& SAM        & 0.49 & 165.54 & 0.37 & 429.25 & 0.24 & 533.19 & 0.06 & 607.70 \\
& SAM~2      & 0.83 & 44.03  & \second{0.81} & 100.95 & \best{0.78} & 104.27 & 0.53 & 210.94 \\
& SAM~3      & \best{0.93} & \best{6.71} & \best{0.82} & 90.20  & \second{0.71} & 109.89 & \best{0.62} & 155.72 \\
& MedSAM     & 0.02 & 89.81  & 0.00 & 183.98 & 0.04 & 162.40 & 0.01 & 148.48 \\
& SAM-Med2D  & 0.68 & 40.67  & 0.63 & \best{73.62} & 0.50 & \second{101.63} & 0.48 & \second{122.43}  \\
& MedSAM2    & 0.77 & 25.96  & 0.77 & \second{75.87} & 0.63 & \best{68.00} & 0.49 & \best{97.05}  \\
& MedSAM3    & \best{0.93} & \second{8.24} & 0.79 & 122.56 & \second{0.71} & 132.67 & \second{0.58} & 158.50 \\
\midrule
\multirow{7}{*}{Invisible (Occluded)}
& SAM        & -- & -- & 0.00 & 498.83 & 0.00 & 587.13 & 0.01 & 616.51 \\
& SAM~2      & -- & -- & 0.00 & 220.57 & 0.00 & 180.08 & 0.00 & 251.46 \\
& SAM~3      & -- & -- & \second{0.09} & \best{80.90} & \second{0.35} & \second{85.45} & 0.52 & 90.40 \\
& MedSAM     & -- & -- & 0.02 & 141.66 & 0.04 & 153.79 & 0.04 & 151.61 \\
& SAM-Med2D  & -- & -- & 0.05 & \second{81.99} & 0.22 & 100.70 & \best{0.67} & \best{83.45}  \\
& MedSAM2    & -- & -- & 0.05 & 181.30 & 0.22 & 108.57 & 0.38 & 110.86 \\
& MedSAM3    & -- & -- & \best{0.18} & 83.47  & \best{0.42} & \best{83.15} & \second{0.66} & \second{86.52}  \\
\bottomrule
\end{tabular}
\end{table}

\begin{table}[H]
\centering
\footnotesize
\caption{Segmentation performance on CVC-300 under Cutout occlusion using Point prompts across clean, low, medium, and high severity levels. \textbf{Bold} and \underline{underline} denote the best and second best per column, respectively.}
\label{tab:sup_cvc300_cutout_point}
\begin{tabular}{llcccccccc}
\toprule
\multirow{2}{*}{Eval Mode} & \multirow{2}{*}{Model} 
& \multicolumn{2}{c}{Clean} 
& \multicolumn{2}{c}{Low} 
& \multicolumn{2}{c}{Medium} 
& \multicolumn{2}{c}{High} \\
\cmidrule(lr){3-4} \cmidrule(lr){5-6} \cmidrule(lr){7-8} \cmidrule(lr){9-10}
& & DSC $\uparrow$ & HD95 $\downarrow$ 
  & DSC $\uparrow$ & HD95 $\downarrow$ 
  & DSC $\uparrow$ & HD95 $\downarrow$ 
  & DSC $\uparrow$ & HD95 $\downarrow$ \\
\midrule
\multirow{7}{*}{Full (Amodal)}
& SAM        & 0.49 & 165.54 & 0.29 & 227.06 & 0.34 & 146.29 & 0.38 & 122.88 \\
& SAM~2      & \second{0.83} & 44.03  & 0.78 & 33.04  & \second{0.54} & 55.09  & 0.52 & 50.47 \\
& SAM~3      & \best{0.93} & \best{6.71} & 0.80 & \second{30.49} & \best{0.75} & \best{41.28} & \best{0.68} & \second{46.37}  \\
& MedSAM     & 0.02 & 89.81  & 0.03 & 81.26  & 0.01 & 82.60  & 0.02 & 72.76 \\
& SAM-Med2D  & 0.68 & 40.67  & 0.54 & 38.21  & 0.45 & \second{45.54} & 0.46 & 51.26 \\
& MedSAM2    & 0.77 & 25.96  & \best{0.83} & \best{20.28} & 0.52 & 71.16  & 0.58 & \best{45.31}  \\
& MedSAM3    & \best{0.93} & \second{8.24} & \second{0.82} & 34.66  & \best{0.75} & 81.03  & \second{0.66} & 73.78 \\
\midrule
\multirow{7}{*}{Visible Only}
& SAM        & 0.49 & 165.54 & 0.29 & 248.05 & 0.28 & 206.11 & 0.06 & 324.66 \\
& SAM~2      & \second{0.83} & 44.03  & \second{0.85} & \second{32.07} & 0.64 & 69.77  & \second{0.69} & 68.06 \\
& SAM~3      & \best{0.93} & \best{6.71} & 0.82 & 37.83  & \second{0.66} & 69.64  & \best{0.78} & \best{33.73}  \\
& MedSAM     & 0.02 & 89.81  & 0.02 & 93.28  & 0.01 & 88.15  & 0.01 & 72.55 \\
& SAM-Med2D  & 0.68 & 40.67  & 0.58 & 35.65  & 0.45 & \best{55.44} & 0.48 & 50.48 \\
& MedSAM2    & 0.77 & 25.96  & \best{0.87} & \best{12.12} & 0.50 & \second{56.18} & 0.59 & \second{41.83}  \\
& MedSAM3    & \best{0.93} & \second{8.24} & 0.80 & 52.75  & \best{0.68} & 82.84  & 0.62 & 78.52 \\
\midrule
\multirow{7}{*}{Invisible (Occluded)}
& SAM        & -- & -- & 0.55 & \second{5.85} & \second{0.60} & 23.91 & \best{0.54} & 50.08 \\
& SAM~2      & -- & -- & 0.18 & 21.06 & 0.51 & 26.73 & 0.48 & 49.48 \\
& SAM~3      & -- & -- & \second{0.64} & \best{5.38} & \second{0.60} & \second{20.38} & 0.50 & 50.49 \\
& MedSAM     & -- & -- & 0.20 & 11.47 & 0.03 & 48.81 & 0.02 & 61.91 \\
& SAM-Med2D  & -- & -- & 0.34 & 5.91  & 0.57 & 24.59 & 0.49 & 47.09 \\
& MedSAM2    & -- & -- & 0.19 & 5.94  & 0.32 & 30.42 & 0.50 & \best{31.28}  \\
& MedSAM3    & -- & -- & \best{0.81} & 7.37  & \best{0.74} & \best{18.44} & \second{0.53} & \second{41.24}  \\
\bottomrule
\end{tabular}
\end{table}

\begin{table}[H]
\centering
\footnotesize
\caption{Segmentation performance on CVC-ColonDB under Surgical Tool occlusion using Box prompts across clean, low, medium, and high severity levels. \textbf{Bold} and \underline{underline} denote the best and second best per column, respectively.}
\label{tab:sup_colondb_tool_box}
\begin{tabular}{llcccccccc}
\toprule
\multirow{2}{*}{Eval Mode} & \multirow{2}{*}{Model} 
& \multicolumn{2}{c}{Clean} 
& \multicolumn{2}{c}{Low} 
& \multicolumn{2}{c}{Medium} 
& \multicolumn{2}{c}{High} \\
\cmidrule(lr){3-4} \cmidrule(lr){5-6} \cmidrule(lr){7-8} \cmidrule(lr){9-10}
& & DSC $\uparrow$ & HD95 $\downarrow$ 
  & DSC $\uparrow$ & HD95 $\downarrow$ 
  & DSC $\uparrow$ & HD95 $\downarrow$ 
  & DSC $\uparrow$ & HD95 $\downarrow$ \\
\midrule
\multirow{7}{*}{Full (Amodal)}
& SAM        & 0.88 & 20.48 & 0.80 & 86.41  & 0.60 & 110.62 & 0.57 & 95.72 \\
& SAM~2      & \second{0.91} & 16.24 & 0.84 & 67.70  & 0.73 & 80.37  & 0.65 & 71.77 \\
& SAM~3      & \best{0.93} & \best{13.51} & \best{0.87} & \second{62.71} & 0.76 & \second{72.79} & 0.63 & 71.57 \\
& MedSAM     & 0.71 & 36.73 & 0.66 & 88.09  & 0.57 & 90.08  & 0.61 & 77.75 \\
& SAM-Med2D  & 0.85 & 28.24 & 0.76 & 77.80  & 0.57 & 97.07  & 0.54 & 79.34 \\
& MedSAM2    & \second{0.91} & 14.73 & \best{0.87} & \best{54.72} & \best{0.82} & \best{62.38} & \best{0.76} & \best{59.00}  \\
& MedSAM3    & \second{0.91} & \second{14.43} & \second{0.85} & 63.42  & \second{0.77} & 73.72  & \second{0.66} & \second{66.89}  \\
\midrule
\multirow{7}{*}{Visible Only}
& SAM        & 0.88 & 20.48 & 0.84 & 62.35  & 0.63 & 73.93  & 0.50 & 67.63 \\
& SAM~2      & \second{0.91} & 16.24 & \second{0.89} & 44.30  & 0.82 & \second{41.57} & \second{0.73} & \second{43.37}  \\
& SAM~3      & \best{0.93} & \best{13.51} & \best{0.91} & \best{35.73} & \best{0.88} & \best{36.50} & \best{0.75} & \best{40.14}  \\
& MedSAM     & 0.71 & 36.73 & 0.66 & 79.30  & 0.46 & 77.80  & 0.28 & 75.33 \\
& SAM-Med2D  & 0.85 & 28.24 & 0.80 & 60.37  & 0.54 & 71.63  & 0.39 & 66.02 \\
& MedSAM2    & \second{0.91} & 14.73 & 0.87 & 46.24  & 0.75 & 65.33  & 0.59 & 64.57 \\
& MedSAM3    & \second{0.91} & \second{14.43} & 0.88 & \second{44.00} & \second{0.84} & 43.42  & 0.72 & 46.67 \\
\midrule
\multirow{7}{*}{Invisible (Occluded)}
& SAM        & -- & -- & 0.05 & 101.92 & 0.29 & 113.93 & 0.50 & 78.48 \\
& SAM~2      & -- & -- & 0.11 & 70.81  & 0.28 & 89.12  & 0.41 & 80.17 \\
& SAM~3      & -- & -- & 0.08 & 65.79  & 0.17 & 82.78  & 0.30 & 73.83 \\
& MedSAM     & -- & -- & \best{0.49} & \second{40.17} & \second{0.64} & \second{50.65} & \best{0.79} & \best{37.08}  \\
& SAM-Med2D  & -- & -- & 0.13 & 73.99  & 0.42 & 78.97  & 0.63 & 56.76 \\
& MedSAM2    & -- & -- & \second{0.48} & \best{36.26} & \best{0.72} & \best{48.13} & \second{0.74} & \second{52.83}  \\
& MedSAM3    & -- & -- & 0.22 & 45.62  & 0.30 & 70.61  & 0.41 & 63.79 \\
\bottomrule
\end{tabular}
\end{table}

\begin{table}[H]
\centering
\footnotesize
\caption{Segmentation performance on CVC-ColonDB under Cutout occlusion using Box prompts across clean, low, medium, and high severity levels. \textbf{Bold} and \underline{underline} denote the best and second best per column, respectively.}
\label{tab:sup_colondb_cutout_box}
\begin{tabular}{llcccccccc}
\toprule
\multirow{2}{*}{Eval Mode} & \multirow{2}{*}{Model} 
& \multicolumn{2}{c}{Clean} 
& \multicolumn{2}{c}{Low} 
& \multicolumn{2}{c}{Medium} 
& \multicolumn{2}{c}{High} \\
\cmidrule(lr){3-4} \cmidrule(lr){5-6} \cmidrule(lr){7-8} \cmidrule(lr){9-10}
& & DSC $\uparrow$ & HD95 $\downarrow$ 
  & DSC $\uparrow$ & HD95 $\downarrow$ 
  & DSC $\uparrow$ & HD95 $\downarrow$ 
  & DSC $\uparrow$ & HD95 $\downarrow$ \\
\midrule
\multirow{7}{*}{Full (Amodal)}
& SAM        & 0.88 & 20.48 & 0.82 & 30.02 & 0.66 & 49.17 & 0.58 & 59.09 \\
& SAM~2      & \second{0.91} & 16.24 & 0.88 & 23.22 & 0.77 & 39.30 & 0.66 & 45.02 \\
& SAM~3      & \best{0.93} & \best{13.51} & \best{0.91} & \best{14.53} & \best{0.90} & \second{29.74} & 0.68 & \best{34.97}  \\
& MedSAM     & 0.71 & 36.73 & 0.65 & 40.02 & 0.58 & 45.92 & 0.65 & 42.88 \\
& SAM-Med2D  & 0.85 & 28.24 & 0.81 & 31.89 & 0.71 & 41.60 & 0.61 & 46.97 \\
& MedSAM2    & \second{0.91} & 14.73 & \second{0.89} & 22.26 & 0.80 & 37.80 & \second{0.72} & 40.93 \\
& MedSAM3    & \second{0.91} & \second{14.43} & \second{0.89} & \second{19.39} & \second{0.86} & \best{28.95} & \best{0.79} & \second{37.34}  \\
\midrule
\multirow{7}{*}{Visible Only}
& SAM        & 0.88 & 20.48 & 0.83 & 21.98 & 0.69 & 29.52 & 0.64 & 30.41 \\
& SAM~2      & \second{0.91} & 16.24 & \best{0.90} & \second{16.72} & \best{0.86} & \best{16.43} & \second{0.84} & \best{17.13}  \\
& SAM~3      & \best{0.93} & \best{13.51} & \second{0.89} & \best{9.59} & 0.76 & 55.54 & \best{0.90} & 45.00 \\
& MedSAM     & 0.71 & 36.73 & 0.59 & 37.32 & 0.40 & 42.58 & 0.31 & 45.25 \\
& SAM-Med2D  & 0.85 & 28.24 & 0.80 & 27.77 & 0.70 & 31.48 & 0.63 & 32.65 \\
& MedSAM2    & \second{0.91} & 14.73 & \second{0.89} & 17.33 & \second{0.84} & \second{21.15} & 0.79 & \second{26.79}  \\
& MedSAM3    & \second{0.91} & \second{14.43} & 0.86 & 23.04 & 0.81 & 27.38 & 0.74 & 33.74 \\
\midrule
\multirow{7}{*}{Invisible (Occluded)}
& SAM        & -- & -- & 0.19 & 17.35 & 0.25 & 32.21 & 0.34 & 38.71 \\
& SAM~2      & -- & -- & 0.23 & 24.89 & 0.18 & 55.05 & 0.21 & 70.73 \\
& SAM~3      & -- & -- & \second{0.68} & \best{6.68} & \best{0.81} & 31.37 & 0.21 & \second{31.11}  \\
& MedSAM     & -- & -- & 0.66 & \second{12.94} & \best{0.81} & \best{18.71} & \best{0.84} & \best{23.03}  \\
& SAM-Med2D  & -- & -- & 0.37 & 18.60 & 0.43 & 39.72 & 0.41 & 48.17 \\
& MedSAM2    & -- & -- & 0.35 & 19.00 & 0.39 & 33.40 & 0.45 & 42.21 \\
& MedSAM3    & -- & -- & \best{0.71} & 13.52 & \second{0.68} & \second{22.44} & \second{0.63} & 33.60 \\
\bottomrule
\end{tabular}
\end{table}

\begin{table}[H]
\centering
\footnotesize
\caption{Segmentation performance on CVC-ColonDB under Surgical Tool occlusion using Point prompts across clean, low, medium, and high severity levels. \textbf{Bold} and \underline{underline} denote the best and second best per column, respectively.}
\label{tab:sup_colondb_tool_point}
\begin{tabular}{llcccccccc}
\toprule
\multirow{2}{*}{Eval Mode} & \multirow{2}{*}{Model} 
& \multicolumn{2}{c}{Clean} 
& \multicolumn{2}{c}{Low} 
& \multicolumn{2}{c}{Medium} 
& \multicolumn{2}{c}{High} \\
\cmidrule(lr){3-4} \cmidrule(lr){5-6} \cmidrule(lr){7-8} \cmidrule(lr){9-10}
& & DSC $\uparrow$ & HD95 $\downarrow$ 
  & DSC $\uparrow$ & HD95 $\downarrow$ 
  & DSC $\uparrow$ & HD95 $\downarrow$ 
  & DSC $\uparrow$ & HD95 $\downarrow$ \\
\midrule
\multirow{7}{*}{Full (Amodal)}
& SAM        & 0.32 & 248.89 & 0.26 & 496.00 & 0.24 & 439.15 & 0.34 & 305.62 \\
& SAM~2      & 0.74 & 72.05  & 0.56 & 241.01 & 0.43 & 248.51 & 0.45 & 177.26 \\
& SAM~3      & \second{0.78} & \second{50.32} & \second{0.65} & \second{132.18} & 0.49 & 170.44 & 0.43 & 155.95 \\
& MedSAM     & 0.02 & 133.46 & 0.02 & 265.20 & 0.02 & 252.97 & 0.02 & 194.77 \\
& SAM-Med2D  & 0.53 & 67.91  & 0.40 & 155.28 & 0.35 & 172.58 & 0.45 & 133.59 \\
& MedSAM2    & 0.73 & 51.94  & 0.64 & \best{114.44} & \second{0.54} & \best{142.81} & \second{0.50} & \second{131.87}  \\
& MedSAM3    & \best{0.81} & \best{47.77} & \best{0.73} & 134.29 & \best{0.56} & \second{170.06} & \best{0.53} & \best{131.18}  \\
\midrule
\multirow{7}{*}{Visible Only}
& SAM        & 0.32 & 248.89 & 0.31 & 476.20 & 0.26 & 507.95 & 0.14 & 583.24 \\
& SAM~2      & 0.74 & 72.05  & 0.65 & 202.48 & 0.58 & 261.97 & 0.53 & 283.68 \\
& SAM~3      & \second{0.78} & \second{50.32} & \second{0.68} & 140.88 & \second{0.64} & 171.77 & \second{0.59} & 189.79 \\
& MedSAM     & 0.02 & 133.46 & 0.02 & 254.89 & 0.02 & 247.95 & 0.01 & 198.21 \\
& SAM-Med2D  & 0.53 & 67.91  & 0.44 & 154.55 & 0.45 & 167.25 & 0.48 & \second{157.12}  \\
& MedSAM2    & 0.73 & 51.94  & 0.67 & \best{109.97} & 0.59 & \best{130.40} & 0.52 & \best{143.27}  \\
& MedSAM3    & \best{0.81} & \best{47.77} & \best{0.74} & \second{136.15} & \best{0.70} & \second{142.54} & \best{0.62} & 158.74 \\
\midrule
\multirow{7}{*}{Invisible (Occluded)}
& SAM        & -- & -- & 0.00 & 630.90 & 0.00 & 586.58 & 0.00 & 620.33 \\
& SAM~2      & -- & -- & 0.00 & 380.16 & 0.01 & 345.82 & 0.00 & 329.40 \\
& SAM~3      & -- & -- & 0.09 & 83.69  & \second{0.33} & 107.34 & 0.40 & 107.33 \\
& MedSAM     & -- & -- & 0.01 & 205.22 & 0.02 & 232.07 & 0.02 & 197.08 \\
& SAM-Med2D  & -- & -- & 0.04 & \best{60.50} & 0.23 & \second{101.50} & \second{0.55} & \best{84.84}  \\
& MedSAM2    & -- & -- & \best{0.19} & \second{72.73} & \best{0.41} & \best{99.85} & \best{0.57} & 109.25 \\
& MedSAM3    & -- & -- & \second{0.16} & 79.03  & 0.31 & 101.84 & 0.53 & \second{104.77}  \\
\bottomrule
\end{tabular}
\end{table}

\begin{table}[H]
\centering
\footnotesize
\caption{Segmentation performance on CVC-ColonDB under Cutout occlusion using Point prompts across clean, low, medium, and high severity levels. \textbf{Bold} and \underline{underline} denote the best and second best per column, respectively.}
\label{tab:sup_colondb_cutout_point}
\begin{tabular}{llcccccccc}
\toprule
\multirow{2}{*}{Eval Mode} & \multirow{2}{*}{Model} 
& \multicolumn{2}{c}{Clean} 
& \multicolumn{2}{c}{Low} 
& \multicolumn{2}{c}{Medium} 
& \multicolumn{2}{c}{High} \\
\cmidrule(lr){3-4} \cmidrule(lr){5-6} \cmidrule(lr){7-8} \cmidrule(lr){9-10}
& & DSC $\uparrow$ & HD95 $\downarrow$ 
  & DSC $\uparrow$ & HD95 $\downarrow$ 
  & DSC $\uparrow$ & HD95 $\downarrow$ 
  & DSC $\uparrow$ & HD95 $\downarrow$ \\
\midrule
\multirow{7}{*}{Full (Amodal)}
& SAM        & 0.32 & 248.89 & 0.27 & 238.48 & 0.29 & 189.87 & 0.46 & 106.36 \\
& SAM~2      & 0.74 & 72.05  & \second{0.66} & 73.48  & 0.55 & 80.03  & 0.51 & \second{89.63}  \\
& SAM~3      & \second{0.78} & \second{50.32} & 0.58 & \best{24.94} & 0.50 & 90.58  & \second{0.58} & 154.00 \\
& MedSAM     & 0.02 & 133.46 & 0.02 & 125.80 & 0.02 & 116.01 & 0.02 & 118.44 \\
& SAM-Med2D  & 0.53 & 67.91  & 0.43 & 71.74  & 0.35 & 79.62  & 0.39 & \best{81.40}  \\
& MedSAM2    & 0.73 & 51.94  & 0.65 & 60.75  & \second{0.58} & \best{65.98} & 0.49 & 105.14 \\
& MedSAM3    & \best{0.81} & \best{47.77} & \best{0.78} & \second{55.53} & \best{0.72} & \second{73.25} & \best{0.63} & 96.59 \\
\midrule
\multirow{7}{*}{Visible Only}
& SAM        & 0.32 & 248.89 & 0.26 & 258.86 & 0.23 & 278.66 & 0.21 & 285.46 \\
& SAM~2      & 0.74 & 72.05  & 0.70 & 75.53  & \second{0.69} & 85.51  & \second{0.66} & 93.40 \\
& SAM~3      & \second{0.78} & \second{50.32} & \best{0.81} & \best{7.47} & 0.43 & 282.65 & \best{0.89} & 158.00 \\
& MedSAM     & 0.02 & 133.46 & 0.02 & 128.57 & 0.02 & 131.78 & 0.01 & 122.71 \\
& SAM-Med2D  & 0.53 & 67.91  & 0.46 & 69.85  & 0.40 & 87.16  & 0.43 & 90.27 \\
& MedSAM2    & 0.73 & 51.94  & 0.67 & \second{61.95} & 0.68 & \best{56.83} & 0.61 & \second{73.46}  \\
& MedSAM3    & \best{0.81} & \best{47.77} & \second{0.74} & 62.34  & \best{0.70} & \second{66.45} & \second{0.66} & \best{73.24}  \\
\midrule
\multirow{7}{*}{Invisible (Occluded)}
& SAM        & -- & -- & 0.43 & 20.41  & 0.41 & 42.18  & \best{0.55} & 56.40 \\
& SAM~2      & -- & -- & 0.11 & 35.37  & 0.27 & 50.86  & \second{0.52} & 59.21 \\
& SAM~3      & -- & -- & \second{0.55} & \second{12.18} & \second{0.42} & \best{14.00} & 0.02 & 138.00 \\
& MedSAM     & -- & -- & 0.04 & 28.53  & 0.04 & 59.64  & 0.02 & 88.42 \\
& SAM-Med2D  & -- & -- & 0.29 & \best{10.88} & 0.38 & \second{32.80} & 0.46 & \second{54.92}  \\
& MedSAM2    & -- & -- & 0.14 & 29.67  & 0.17 & 45.13  & 0.40 & 56.12 \\
& MedSAM3    & -- & -- & \best{0.57} & 17.42  & \best{0.52} & 35.73  & \best{0.55} & \best{52.44}  \\
\bottomrule
\end{tabular}
\end{table}

\begin{table}[H]
\centering
\footnotesize
\caption{Segmentation performance on ETIS-LaribPolypDB under Surgical Tool occlusion using Box prompts across clean, low, medium, and high severity levels. \textbf{Bold} and \underline{underline} denote the best and second best per column, respectively.}
\label{tab:sup_etis_tool_box}
\begin{tabular}{llcccccccc}
\toprule
\multirow{2}{*}{Eval Mode} & \multirow{2}{*}{Model} 
& \multicolumn{2}{c}{Clean} 
& \multicolumn{2}{c}{Low} 
& \multicolumn{2}{c}{Medium} 
& \multicolumn{2}{c}{High} \\
\cmidrule(lr){3-4} \cmidrule(lr){5-6} \cmidrule(lr){7-8} \cmidrule(lr){9-10}
& & DSC $\uparrow$ & HD95 $\downarrow$ 
  & DSC $\uparrow$ & HD95 $\downarrow$ 
  & DSC $\uparrow$ & HD95 $\downarrow$ 
  & DSC $\uparrow$ & HD95 $\downarrow$ \\
\midrule
\multirow{7}{*}{Full (Amodal)}
& SAM        & 0.91 & 33.13 & 0.81 & 45.30  & 0.66 & 84.53  & 0.55 & 91.15 \\
& SAM~2      & 0.91 & 33.18 & \second{0.84} & 48.94  & 0.74 & 70.69  & 0.62 & 82.04 \\
& SAM~3      & \best{0.94} & \best{22.15} & \second{0.84} & \best{32.71} & \second{0.76} & \second{66.51} & 0.60 & \second{66.48}  \\
& MedSAM     & 0.78 & 47.40 & 0.72 & 46.89  & 0.62 & 71.77  & 0.59 & 76.70 \\
& SAM-Med2D  & 0.84 & 52.78 & 0.78 & 46.87  & 0.61 & 77.94  & 0.50 & 87.52 \\
& MedSAM2    & 0.91 & 30.57 & \best{0.87} & \second{34.29} & \best{0.81} & \best{51.99} & \best{0.77} & \best{56.57}  \\
& MedSAM3    & \second{0.92} & \second{29.13} & \best{0.87} & 34.55  & \second{0.76} & 69.65  & \second{0.68} & 70.43 \\
\midrule
\multirow{7}{*}{Visible Only}
& SAM        & 0.91 & 33.13 & 0.84 & 38.60  & 0.78 & 47.37  & 0.55 & 59.78 \\
& SAM~2      & 0.91 & 33.18 & 0.88 & 36.96  & \second{0.86} & \second{33.84} & 0.72 & \second{49.82}  \\
& SAM~3      & \best{0.94} & \best{22.15} & \second{0.89} & \best{18.82} & \best{0.89} & \best{26.92} & \best{0.80} & \best{29.95}  \\
& MedSAM     & 0.78 & 47.40 & 0.76 & 39.70  & 0.60 & 53.23  & 0.39 & 66.79 \\
& SAM-Med2D  & 0.84 & 52.78 & 0.82 & 36.60  & 0.65 & 52.55  & 0.49 & 63.00 \\
& MedSAM2    & 0.91 & 30.57 & \second{0.89} & 29.51  & 0.80 & 48.56  & 0.64 & 64.14 \\
& MedSAM3    & \second{0.92} & \second{29.13} & \best{0.91} & \second{22.60} & 0.84 & 42.67  & \second{0.74} & 50.56 \\
\midrule
\multirow{7}{*}{Invisible (Occluded)}
& SAM        & -- & -- & 0.03 & 75.04  & 0.16 & 115.42 & 0.44 & 93.66 \\
& SAM~2      & -- & -- & 0.12 & 47.62  & 0.16 & 73.67  & 0.39 & 75.98 \\
& SAM~3      & -- & -- & 0.04 & 51.98  & 0.11 & 75.52  & 0.18 & 76.34 \\
& MedSAM     & -- & -- & \second{0.22} & 43.00  & \second{0.55} & \best{41.49} & \second{0.69} & \best{48.01}  \\
& SAM-Med2D  & -- & -- & 0.10 & 54.94  & 0.30 & 65.36  & 0.48 & 67.48 \\
& MedSAM2    & -- & -- & \best{0.36} & \best{29.42} & \best{0.56} & \second{43.86} & \best{0.71} & \second{51.71}  \\
& MedSAM3    & -- & -- & 0.15 & \second{33.02} & 0.27 & 63.41  & 0.44 & 63.85 \\
\bottomrule
\end{tabular}
\end{table}

\begin{table}[H]
\centering
\footnotesize
\caption{Segmentation performance on ETIS-LaribPolypDB under Cutout occlusion using Box prompts across clean, low, medium, and high severity levels. \textbf{Bold} and \underline{underline} denote the best and second best per column, respectively.}
\label{tab:sup_etis_cutout_box}
\begin{tabular}{llcccccccc}
\toprule
\multirow{2}{*}{Eval Mode} & \multirow{2}{*}{Model} 
& \multicolumn{2}{c}{Clean} 
& \multicolumn{2}{c}{Low} 
& \multicolumn{2}{c}{Medium} 
& \multicolumn{2}{c}{High} \\
\cmidrule(lr){3-4} \cmidrule(lr){5-6} \cmidrule(lr){7-8} \cmidrule(lr){9-10}
& & DSC $\uparrow$ & HD95 $\downarrow$ 
  & DSC $\uparrow$ & HD95 $\downarrow$ 
  & DSC $\uparrow$ & HD95 $\downarrow$ 
  & DSC $\uparrow$ & HD95 $\downarrow$ \\
\midrule
\multirow{7}{*}{Full (Amodal)}
& SAM        & 0.91 & 33.13 & 0.82 & 50.09  & 0.68 & 79.46  & 0.63 & 92.08 \\
& SAM~2      & 0.91 & 33.18 & 0.88 & 43.65  & 0.75 & 70.49  & 0.66 & 79.74 \\
& SAM~3      & \best{0.94} & \best{22.15} & \best{0.92} & \best{31.75} & \best{0.86} & \second{60.73} & \best{0.77} & 69.34 \\
& MedSAM     & 0.78 & 47.40 & 0.70 & 55.99  & 0.67 & 69.33  & 0.67 & \second{68.51}  \\
& SAM-Med2D  & 0.84 & 52.78 & 0.79 & 57.14  & 0.72 & 71.58  & 0.62 & 81.02 \\
& MedSAM2    & 0.91 & 30.57 & 0.87 & 42.51  & \second{0.80} & 66.70  & 0.67 & 80.92 \\
& MedSAM3    & \second{0.92} & \second{29.13} & \second{0.91} & \second{31.91} & \best{0.86} & \best{57.10} & \second{0.76} & \best{64.38}  \\
\midrule
\multirow{7}{*}{Visible Only}
& SAM        & 0.91 & 33.13 & 0.82 & 39.68  & 0.67 & 47.75  & 0.70 & 44.71 \\
& SAM~2      & 0.91 & 33.18 & \best{0.90} & \second{33.61} & \second{0.84} & \second{37.76} & \best{0.81} & \best{38.29}  \\
& SAM~3      & \best{0.94} & \best{22.15} & \best{0.90} & \best{28.91} & 0.83 & 43.71  & \best{0.81} & \second{41.42}  \\
& MedSAM     & 0.78 & 47.40 & 0.66 & 53.10  & 0.50 & 67.51  & 0.37 & 76.33 \\
& SAM-Med2D  & 0.84 & 52.78 & 0.78 & 50.28  & 0.68 & 59.32  & 0.64 & 56.64 \\
& MedSAM2    & 0.91 & 30.57 & \second{0.88} & 34.04  & \best{0.87} & \best{31.47} & \second{0.79} & 49.93 \\
& MedSAM3    & \second{0.92} & \second{29.13} & \second{0.88} & 37.20  & 0.80 & 52.04  & 0.77 & 46.84 \\
\midrule
\multirow{7}{*}{Invisible (Occluded)}
& SAM        & -- & -- & 0.16 & 38.30  & 0.29 & 61.02  & 0.33 & 84.98 \\
& SAM~2      & -- & -- & 0.21 & 42.73  & 0.22 & 90.07  & 0.27 & 108.33 \\
& SAM~3      & -- & -- & 0.57 & \best{19.87} & 0.63 & 46.34  & 0.52 & \second{52.22}  \\
& MedSAM     & -- & -- & \second{0.59} & 24.60  & \best{0.82} & \best{37.02} & \best{0.84} & \best{37.10}  \\
& SAM-Med2D  & -- & -- & 0.41 & 28.75  & 0.51 & 61.39  & 0.39 & 75.26 \\
& MedSAM2    & -- & -- & 0.27 & 30.30  & 0.28 & 63.97  & 0.31 & 82.63 \\
& MedSAM3    & -- & -- & \best{0.67} & \second{21.27} & \second{0.67} & \second{44.66} & \second{0.57} & 52.43 \\
\bottomrule
\end{tabular}
\end{table}

\begin{table}[H]
\centering
\footnotesize
\caption{Segmentation performance on ETIS-LaribPolypDB under Surgical Tool occlusion using Point prompts across clean, low, medium, and high severity levels. \textbf{Bold} and \underline{underline} denote the best and second best per column, respectively.}
\label{tab:sup_etis_tool_point}
\begin{tabular}{llcccccccc}
\toprule
\multirow{2}{*}{Eval Mode} & \multirow{2}{*}{Model} 
& \multicolumn{2}{c}{Clean} 
& \multicolumn{2}{c}{Low} 
& \multicolumn{2}{c}{Medium} 
& \multicolumn{2}{c}{High} \\
\cmidrule(lr){3-4} \cmidrule(lr){5-6} \cmidrule(lr){7-8} \cmidrule(lr){9-10}
& & DSC $\uparrow$ & HD95 $\downarrow$ 
  & DSC $\uparrow$ & HD95 $\downarrow$ 
  & DSC $\uparrow$ & HD95 $\downarrow$ 
  & DSC $\uparrow$ & HD95 $\downarrow$ \\
\midrule
\multirow{7}{*}{Full (Amodal)}
& SAM        & 0.46 & 405.77 & 0.28 & 503.74 & 0.28 & 414.99 & 0.32 & 296.23 \\
& SAM~2      & 0.69 & 225.01 & 0.57 & 220.11 & 0.44 & 245.45 & 0.40 & 201.71 \\
& SAM~3      & \second{0.75} & 125.54 & \second{0.66} & 133.30 & \second{0.48} & 161.10 & \second{0.44} & 164.78 \\
& MedSAM     & 0.02 & 232.68 & 0.03 & 180.45 & 0.03 & 175.92 & 0.02 & 192.56 \\
& SAM-Med2D  & 0.67 & \best{112.44} & 0.58 & \best{102.98} & 0.42 & \second{132.24} & \second{0.44} & \best{134.33}  \\
& MedSAM2    & 0.71 & \second{123.13} & 0.61 & \second{125.50} & \best{0.56} & \best{115.28} & \best{0.47} & \second{140.05}  \\
& MedSAM3    & \best{0.79} & 142.88 & \best{0.70} & 155.37 & \best{0.56} & 152.14 & \best{0.47} & 146.07 \\
\midrule
\multirow{7}{*}{Visible Only}
& SAM        & 0.46 & 405.77 & 0.35 & 457.31 & 0.37 & 442.35 & 0.27 & 498.70 \\
& SAM~2      & 0.69 & 225.01 & 0.66 & 203.64 & 0.65 & 214.55 & 0.56 & 287.09 \\
& SAM~3      & \second{0.75} & 125.54 & \best{0.76} & \second{113.28} & \second{0.66} & \second{119.13} & \second{0.57} & 197.49 \\
& MedSAM     & 0.02 & 232.68 & 0.02 & 181.17 & 0.02 & 203.16 & 0.03 & 179.35 \\
& SAM-Med2D  & 0.67 & \best{112.44} & 0.65 & \best{81.94} & 0.55 & 132.59 & 0.55 & \second{165.54}  \\
& MedSAM2    & 0.71 & \second{123.13} & 0.63 & 117.25 & 0.64 & \best{96.70} & 0.51 & \best{142.46}  \\
& MedSAM3    & \best{0.79} & 142.88 & \second{0.73} & 153.81 & \best{0.69} & 154.72 & \best{0.58} & 216.75 \\
\midrule
\multirow{7}{*}{Invisible (Occluded)}
& SAM        & -- & -- & 0.00 & 546.40 & 0.00 & 514.11 & 0.00 & 541.68 \\
& SAM~2      & -- & -- & 0.00 & 292.00 & 0.00 & 301.52 & 0.00 & 320.93 \\
& SAM~3      & -- & -- & 0.05 & \second{83.38} & 0.21 & 110.88 & 0.35 & 104.73 \\
& MedSAM     & -- & -- & 0.00 & 164.57 & 0.02 & 189.95 & 0.03 & 170.84 \\
& SAM-Med2D  & -- & -- & \second{0.06} & 83.73  & \second{0.22} & \best{94.73} & \best{0.52} & \second{101.17}  \\
& MedSAM2    & -- & -- & 0.02 & 210.20 & 0.11 & 158.52 & 0.15 & 167.37 \\
& MedSAM3    & -- & -- & \best{0.13} & \best{80.06} & \best{0.24} & \second{95.62} & \second{0.47} & \best{101.12}  \\
\bottomrule
\end{tabular}
\end{table}

\begin{table}[H]
\centering
\footnotesize
\caption{Segmentation performance on ETIS-LaribPolypDB under Cutout occlusion using Point prompts across clean, low, medium, and high severity levels. \textbf{Bold} and \underline{underline} denote the best and second best per column, respectively.}
\label{tab:sup_etis_cutout_point}
\begin{tabular}{llcccccccc}
\toprule
\multirow{2}{*}{Eval Mode} & \multirow{2}{*}{Model} 
& \multicolumn{2}{c}{Clean} 
& \multicolumn{2}{c}{Low} 
& \multicolumn{2}{c}{Medium} 
& \multicolumn{2}{c}{High} \\
\cmidrule(lr){3-4} \cmidrule(lr){5-6} \cmidrule(lr){7-8} \cmidrule(lr){9-10}
& & DSC $\uparrow$ & HD95 $\downarrow$ 
  & DSC $\uparrow$ & HD95 $\downarrow$ 
  & DSC $\uparrow$ & HD95 $\downarrow$ 
  & DSC $\uparrow$ & HD95 $\downarrow$ \\
\midrule
\multirow{7}{*}{Full (Amodal)}
& SAM        & 0.46 & 405.77 & 0.40 & 382.44 & 0.32 & 377.71 & 0.45 & 209.90 \\
& SAM~2      & 0.69 & 225.01 & 0.65 & 165.76 & 0.48 & 193.81 & 0.49 & 188.75 \\
& SAM~3      & \second{0.75} & 125.54 & \second{0.78} & \best{98.36} & \best{0.64} & 140.81 & \best{0.61} & \second{153.30}  \\
& MedSAM     & 0.02 & 232.68 & 0.03 & 198.13 & 0.03 & 207.02 & 0.02 & 226.62 \\
& SAM-Med2D  & 0.67 & \best{112.44} & 0.58 & \second{103.03} & 0.46 & \second{139.07} & 0.49 & \best{136.29}  \\
& MedSAM2    & 0.71 & \second{123.13} & 0.66 & 122.91 & \second{0.54} & \best{121.36} & 0.41 & 222.15 \\
& MedSAM3    & \best{0.79} & 142.88 & \best{0.79} & 133.93 & \best{0.64} & 215.01 & \second{0.53} & 247.12 \\
\midrule
\multirow{7}{*}{Visible Only}
& SAM        & 0.46 & 405.77 & 0.34 & 463.27 & 0.33 & 532.38 & 0.37 & 454.53 \\
& SAM~2      & 0.69 & 225.01 & 0.67 & 201.99 & 0.63 & 235.22 & \second{0.63} & 240.06 \\
& SAM~3      & \second{0.75} & 125.54 & \best{0.79} & \best{102.31} & \best{0.74} & \second{130.19} & \best{0.71} & \best{120.16}  \\
& MedSAM     & 0.02 & 232.68 & 0.02 & 210.44 & 0.02 & 202.64 & 0.01 & 204.14 \\
& SAM-Med2D  & 0.67 & \best{112.44} & 0.60 & \second{104.23} & 0.52 & 138.13 & 0.52 & \second{139.44}  \\
& MedSAM2    & 0.71 & \second{123.13} & 0.70 & 114.31 & \second{0.68} & \best{91.83} & 0.59 & 179.20 \\
& MedSAM3    & \best{0.79} & 142.88 & \second{0.76} & 124.01 & 0.63 & 228.50 & 0.60 & 194.87 \\
\midrule
\multirow{7}{*}{Invisible (Occluded)}
& SAM        & -- & -- & 0.28 & 48.86  & 0.30 & 80.21  & 0.44 & 97.80 \\
& SAM~2      & -- & -- & 0.13 & 51.98  & 0.27 & 93.70  & 0.41 & 120.41 \\
& SAM~3      & -- & -- & \second{0.42} & 28.96  & 0.39 & 65.40  & \second{0.45} & 95.65 \\
& MedSAM     & -- & -- & 0.04 & 56.35  & 0.07 & 103.81 & 0.02 & 126.89 \\
& SAM-Med2D  & -- & -- & 0.39 & \best{17.24} & \second{0.50} & \second{62.61} & \best{0.47} & \best{86.63}  \\
& MedSAM2    & -- & -- & 0.08 & 44.50  & 0.12 & 84.81  & 0.23 & 105.53 \\
& MedSAM3    & -- & -- & \best{0.59} & \second{28.54} & \best{0.54} & \best{52.76} & \second{0.45} & \second{90.78}  \\
\bottomrule
\end{tabular}
\end{table}

\end{document}